\documentclass[conference]{IEEEtran}
\IEEEoverridecommandlockouts
\usepackage{tikz}
\usepackage{cite}
\usepackage{amsmath,amssymb,amsfonts}
\usepackage{algorithmic}
\usepackage{graphicx}
\usepackage{textcomp}
\usepackage{xcolor}
\usepackage{float}
\usepackage[small]{caption}
\usepackage{subcaption}
\usepackage{placeins}     
\usepackage{epsfig}
\usepackage{svg}
\usepackage[super]{nth}
\usepackage{dblfloatfix}
\usepackage[autostyle]{csquotes} 

\newcommand\copyrighttext{%
  \footnotesize \textcopyright 2020 IEEE. Personal use of this material is permitted.
  Permission from IEEE must be obtained for all other uses, in any current or future
  media, including reprinting/republishing this material for advertising or promotional
  purposes, creating new collective works, for resale or redistribution to servers or
  lists, or reuse of any copyrighted component of this work in other works.}
\newcommand\copyrightnotice{%
\begin{tikzpicture}[remember picture,overlay]
\node[anchor=south,yshift=10pt] at (current page.south) {\fbox{\parbox{\dimexpr\textwidth-\fboxsep-\fboxrule\relax}{\copyrighttext}}};
\end{tikzpicture}%
}
\newcommand{\overbar}[1]{\mkern 1.5mu\overline{\mkern-1.5mu#1\mkern-1.5mu}\mkern 1.5mu}
\def\BibTeX{{\rm B\kern-.05em{\sc i\kern-.025em b}\kern-.08em
    T\kern-.1667em\lower.7ex\hbox{E}\kern-.125emX}}
\begin{document}

\title{Separating the Effects of Batch Normalization on CNN Training Speed and Stability Using Classical Adaptive Filter Theory
}

\author{\IEEEauthorblockN{Elaina Chai\IEEEauthorrefmark{1}, Mert Pilanci\IEEEauthorrefmark{2}, Boris Murmann\IEEEauthorrefmark{3}}
\IEEEauthorblockA{\textit{Dept. of Electrical Engineering}, \textit{Stanford University}\\
Stanford, CA \\
Email: \IEEEauthorrefmark{1}echai@stanford.edu, \IEEEauthorrefmark{2}pilanci@stanford.edu, \IEEEauthorrefmark{3}murmann@stanford.edu}
}

\maketitle
\copyrightnotice
\begin{abstract}
Batch Normalization (BatchNorm) is commonly used in Convolutional Neural Networks (CNNs) to improve training speed and stability. However, there is still limited consensus on why this technique is effective. This paper uses concepts from the traditional adaptive filter domain to provide insight into the dynamics and inner workings of BatchNorm. First, we show that the convolution weight updates have natural modes whose stability and convergence speed are tied to the eigenvalues of the input autocorrelation matrices, which are controlled by BatchNorm through the convolution layers’ channel-wise structure. Furthermore, our experiments demonstrate that the speed and stability benefits are distinct effects. At low learning rates, it is BatchNorm’s amplification of the smallest eigenvalues that improves convergence speed, while at high learning rates, it is BatchNorm’s suppression of the largest eigenvalues that ensures stability. Lastly, we prove that in the first training step, when normalization is needed most, BatchNorm satisfies the same optimization as Normalized Least Mean Square (NLMS), while it continues to approximate this condition in subsequent steps. The analyses provided in this paper lay the groundwork for gaining further insight into the operation of modern neural network structures using adaptive filter theory.
\end{abstract}


\section{Introduction}

The Deep Neural Network (DNN) community is split into two groups. One group drives progress empirically and has delivered innovations in new network architectures such as Residual Networks (ResNets) \cite{He2016} and training techniques such as Batch Normalization (BatchNorm) \cite{ioffe2015batch}. The other group focuses on understanding why these contributions work \cite{Hanin2018} \cite{yang2019MeanFieldTheoryBatchNorm}. However, experiment-driven research has become the prevailing paradigm due to widespread access to large datasets and GPU-accelerated frameworks \cite{PyTorchNEURIPS2019_9015}, resulting in a growing gap between DNN innovations and their theoretical understanding.

An example of where this gap exists is BatchNorm, a channel-wise normalization technique popular in CNN training. The primary benefit is in maintaining network stability at higher learning rates. At lower learning rates, BatchNorm increases training speed. In networks such as ResNet \cite{He2016}, BatchNorm is critical to convergence. The creators of BatchNorm theorized that BatchNorm reduces the effect of Internal Covariate Shift (ICS) \cite{ioffe2015batch}. Later publications have since disproved the impact of ICS. For example, the authors of \cite{santurkar2018does} demonstrated that they could deliberately induce ICS without affecting convergence speed. Works such as \cite{Zhang2019} and \cite{Balduzzi2017} experimentally demonstrated that BatchNorm helps control exploding gradients but ultimately provide conflicting hypotheses. These works show that despite BatchNorm's popularity, there is still a debate on why BatchNorm helps training.


This work focuses on improving our theoretical understanding of BatchNorm. Unlike prior work in this area, we leverage insight from the traditional adaptive filter domain. Although modern DNNs are considered a separate field, there are reasons to expect similarities between classical adaptive filter theory and modern DNN techniques. Neural networks originated from the signal processing community in the early 1960s \cite{widrow1960adaline} and were treated as a particular form of an adaptive filter. Like modern DNNs, adaptive filters are trained using gradient descent to minimize a loss function such as Least Mean Squares (LMS) \cite{widrow1960adaptive}. Additionally, BatchNorm bears a strong resemblance to Normalized LMS (NLMS), a variant of LMS that reduces its sensitivity to the learning rate and increases convergence speed. Thus, it is reasonable to expect that classical adaptive filters can provide insight into understanding BatchNorm. We explore the similarities and differences in this paper. More concretely, this work:

\begin{itemize}
    \item restructures CNNs to follow the traditional adaptive filter notation (Section \ref{section: Restructure}). We address the handling of CNN nonlinearities and the global cost function over multiple layers. This explicit recasting is necessary to reconcile the differences between CNNs and adaptive filters. 
    \item demonstrates that CNNs, similar to adaptive filters, have natural modes, stability bounds, and training speeds that are controlled by the input autocorrelation matrix's eigenvalues (Section \ref{section: Channel_wise_Bounds}). We demonstrate how the sliding convolution mechanisms create uniform statistics within channels, leading to pseudo-whitening effects via channel-wise normalization. This results in better-conditioned eigenvalues. 
    \item  analyzes two BatchNorm variants to explore their impact on the eigenvalues. We show how the pseudo-whitening effects lead to improved bounds and study the effects on training speed and network stability (Section \ref{section: Experiment_NaturalModes}).
    \item proves that the Principle of Minimum Disturbance (PMD) can be applied to CNNs and that under certain conditions, BatchNorm placed before the convolution operation is equivalent to NLMS (Section \ref{section: NLMS}).
\end{itemize}

\section{Casting CNN Features as Adaptive Filters}\label{section: Restructure}



\subsection{Adaptive Filter Definitions}\label{Traditional_LMS}

\subsubsection{Variables} At time step \(n\) for adaptive filters with vector weights and scalar outputs, we use the following variables:

\begin{itemize}
    \item \(\mathbb{X}(n)\): column vector input 
    \item \(\mathbb{W}(n)\): column vector weight
    \item \(d(n) = \mathbb{W}^T(n+1) \mathbb{X}(n)\): scalar desired response 
    \item \(y(n)= \mathbb{X}^T(n) \mathbb{W}(n) = \mathbb{W}^T(n) \mathbb{X}(n)\): scalar output 
    \item \(\varepsilon(n) = d(n)-y(n) = d(n) - \mathbb{W}^T(n)\mathbb{X}(n)\): scalar error 
    \item \(\delta_\mathbb{W}(n+1) = \mathbb{W}(n+1) -\mathbb{W}(n)\)
\end{itemize}

\subsubsection{Convolution}
We use \textit{convolution} to refer to the CNN's spatial convolutions instead of the temporal convolution typical in signal processing.

\subsection{The Convolution Layer as an Adaptive Filter} \label{section:Recast_CNN}

\subsubsection{Restructuring the CNN Convolution Layer}

For CNNs with image inputs, tensors are the default structures for the weights, inputs, and outputs, with the following dimensions:

\begin{itemize}
    \item Weight dimensions input channel \(\times\) output channel \(\times\) height \(\times\) width: \(IC \times OC \times H \times W\)
    \item Input activations input height \(\times\) input width \(\times\) input channels: \(IH \times IW \times IC\)
    \item Output activations output height \(\times\) output width \(\times\) output channels: \(OH \times OW \times OC\)
\end{itemize}

\begin{figure}[t]
\centering
\begin{subfigure}{0.24\textwidth}
  \centering
  \includegraphics[width=1.0\linewidth]{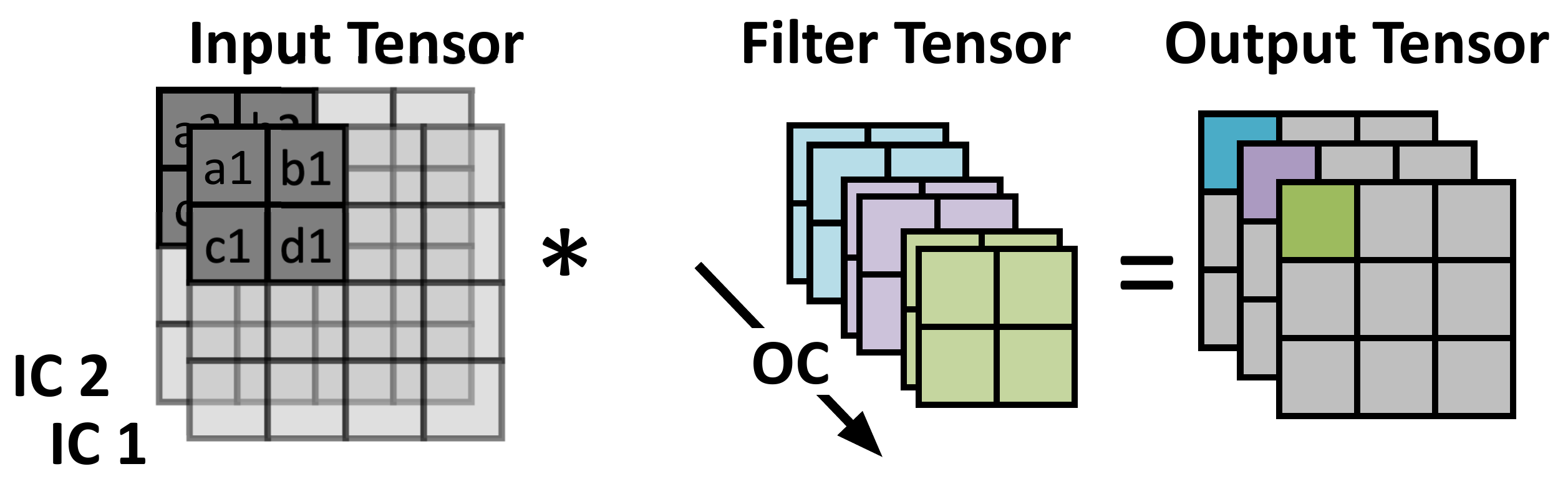}
  \caption{Original CNN spatial convolution}
  \label{fig:Before_Restructure_Conv_IC2_OC3}
\end{subfigure}
\begin{subfigure}{0.24\textwidth}
  \centering
  \includegraphics[width=1.0\linewidth]{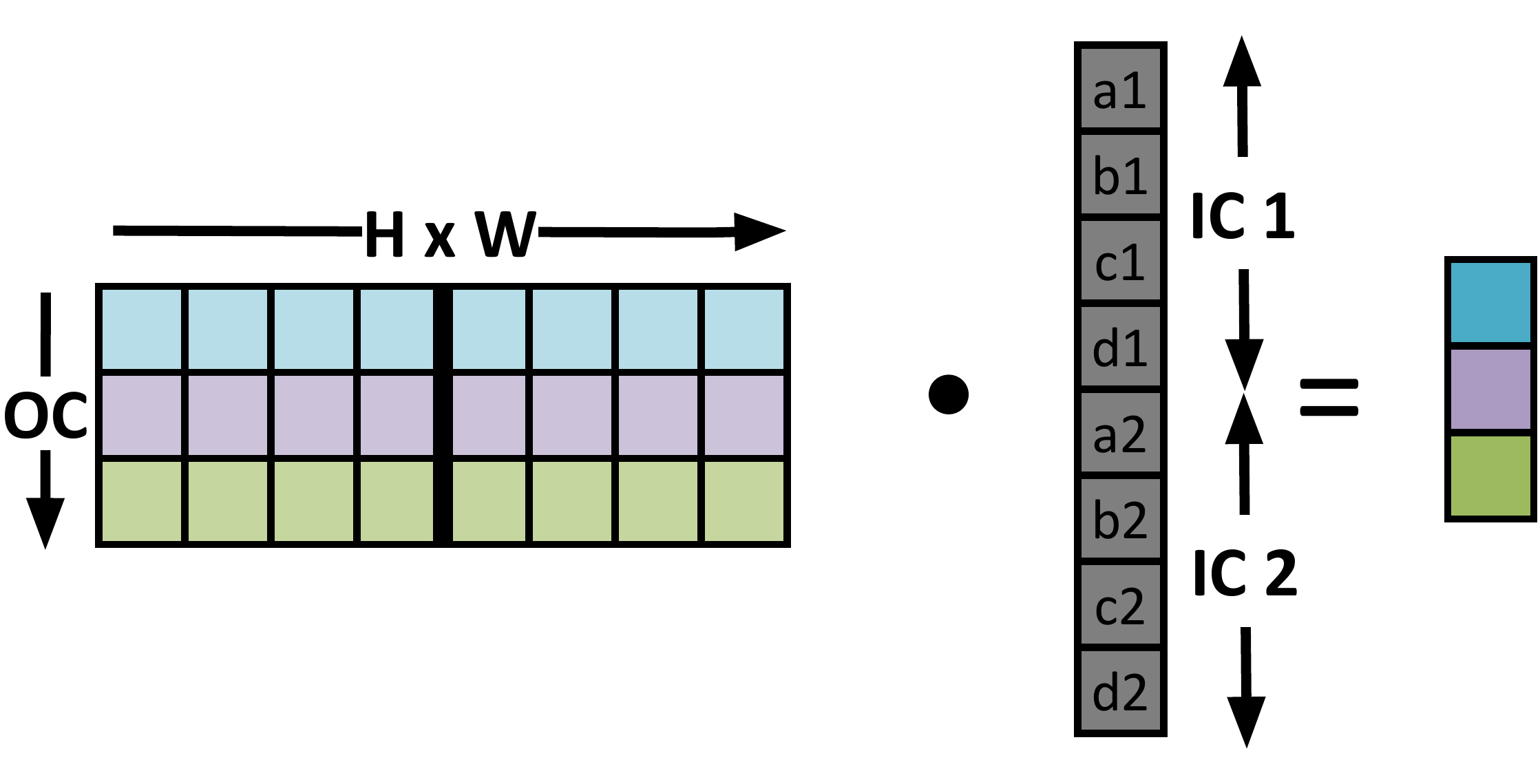}
  \caption{CNN convolution restructured as a matrix-vector product}
  \label{fig:After_Restructure_Conv_IC2_OC3}
\end{subfigure}
\caption{Restructuring the multi-channel spatial convolution as a dot-product, \(IC=2\) and \(OC=3\).}
\label{fig:Restructure_Conv_IC2_OC3}
\end{figure}

We restructure the 4D spatial convolution into a matrix-vector multiplication, as shown in Fig. \ref{fig:Restructure_Conv_IC2_OC3}. Without loss of generality, we restrict the analysis to a single output channel and single batch because operations along the output channels and batches are independent of one another. For brevity, we drop references to the time step \(n\).

\subsubsection{Variables}
Given this restructured form, we adapt the framework from Section \ref{Traditional_LMS} to describe the CNN layer components for a given convolution layer \(l\): 

\begin{itemize}
    \item \(\mathbb{W}^{(l)}\): filter weight vector of size \(IC\times H\times W\). \(w_k^{(l)}\) is the \(k\)\textsuperscript{th} element in \(\mathbb{W}^{(l)}\). When there are multiple output channels, the weight array is the matrix \(\mathbf{W}^{(l)}\).
    \item \(\mathbb{E}^{(l)}\): unrolled local error vector of length \(M\). \(e^{(m,l)}\) is the \(m\)\textsuperscript{th} element in \(\mathbb{E}^{(l)}\). 
    \item \(\mathbb{D}^{(l)}\): unrolled local desired response vector of length \(M\). \(d^{(m,l)}\) is the \(m\)\textsuperscript{th} element in \(\mathbb{D}^{(l)}\). 
    \item \(\mathbb{X}^{(m,l)}\): input vector of size \(IC \times H \times W\). It is the unrolled patch of the spatial input map that is convolved with the filter. \(x_k^{(m,l)}\) is the \(k\)\textsuperscript{th} element in this vector. 
    \item \(\mathbf{X}^{(l)}\): input matrix of size \((IC\times H\times W) \times M\). It is the input when \(\mathbb{X}^{(m,l)}\) is expanded along the spatial convolution strides. Each stride, indexed by \(m\), corresponds to a vector \(\mathbb{X}^{(m,l)}\) and a new output pixel.
    \item \(\mathbb{Y}^{(l)} = \mathbf{X}^{(l)T}\mathbb{W}^{(l)}\): unrolled output vector of length \(M\). \(y^{(m,l)} =\mathbb{W}^{(l)T}\mathbb{X}^{(m,l)}\) is the \(m\)\textsuperscript{th} element in \(\mathbb{Y}^{(l)}\).

\end{itemize}

\begin{figure}[t]
\centering
\begin{subfigure}{0.2\textwidth}
  \centering
  \includegraphics[width=1\linewidth]{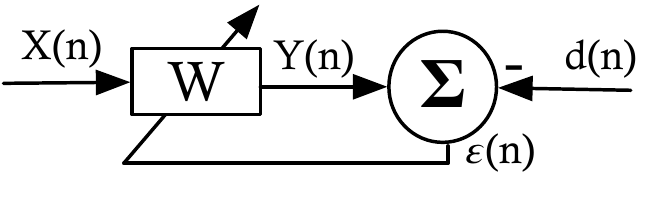}
  \caption{Generic adaptive filter}
  \label{fig:generic_adaptive_filter}
\end{subfigure}
\begin{subfigure}{0.27\textwidth}
  \centering
  \includegraphics[width=1\linewidth]{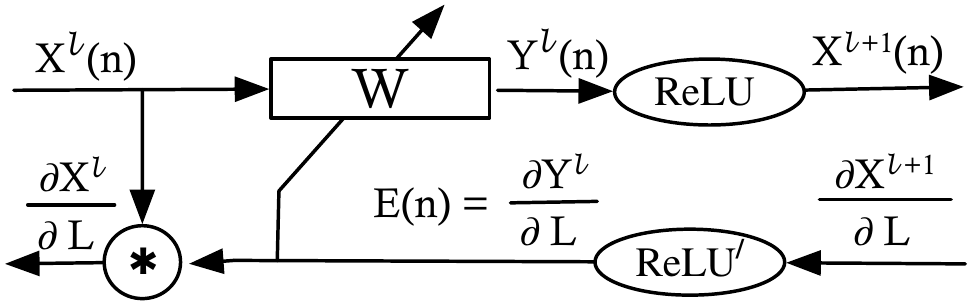}
  \caption{Convolution layer weight update}
  \label{fig:recast_conv_layer_adaptive}
\end{subfigure}
\caption{Convolution layer recast as an adaptive filter.}
\label{fig:Recast_CNN}
\end{figure}

As in the adaptive filter format, the goal of CNN training is to update \(\mathbb{W}^{(l)}\) for all \(l\) so that the desired response \(\mathbb{D}^{(l)}\) eventually matches the local error vector \(\mathbb{E}^{(l)}\). Fig. \ref{fig:Recast_CNN} shows the recast convolution layer with these terms, the ReLU nonlinearity, and the backpropagation path.

\subsubsection{The Restructured Loss Function}

To make the natural modes analysis of Section \ref{section: Natural Modes} tractable, we transform the global loss function \(\nabla J_\theta\) (defined over the entire network architecture) into \(\nabla_{\mathbb{W}^{(l)}}J\), which is a function of only terms available in the current layer \(l\). We start with \(J_\theta\) as a function of only the layer input \(\mathbb{X}^{(l)}\), the layer weights \(\mathbb{W}^{(l)}\) and the downstream weights \(\mathbb{W}^{(l+i)}\) (where \(i > 1\)). At time step \(n\), since the downstream updates do not affect the update at layer \(l\) until time step \(n+1\), we safely fix all downstream weights as constants, leaving \(\mathbb{X}^{(l)}\) and \(\mathbb{W}^{(l)}\) as the only variables in \(\nabla J_\theta\). \(\nabla J_\theta = \nabla_{\mathbb{W}^{(l)}}J\), where \(\nabla_{\mathbb{W}^{(l)}}J\) is the local gradient calculated using the chain rule. For the \(k\)\textsuperscript{th} element in \(\nabla J\):

\begin{align} \label{eq:local_gradient_basic}
\nabla_{\mathbb{W}^{l}_k}J &= \sum_{m=0}^{M-1}\frac{\partial y^{(m,l)}}{\partial w_k^{(l)}}  \frac{\partial J}{\partial y^{(m,l)}} = \sum_{m=0}^{M-1} x^{(m,l)}_{k} e^{(m,l)} = \overrightarrow{\mathbf{X}}_{k,*}^{(l)} \mathbb{E}^{(l)}
\end{align}

\noindent \(\overrightarrow{\mathbf{X}}_{k,*}^{(l)} = [x^{(0,l)}_{k}\; x^{(1,l)}_{k}\; ...\;x^{(M-1,l)}_{k}]\) is the \(k\)\textsuperscript{th} row of \(\mathbf{X}^{(l)}\). We now have an expression for \(\nabla J\) that is usable in the derivation of the natural modes.

\section{Natural Modes and Normalization Effects}\label{section: Channel_wise_Bounds}

This section shows that the restructured CNN layer has natural modes similar to adaptive filters \cite{Haykin:2002}\cite{widrow1971adaptive}. 

\subsection{Derivation of the Natural Modes} \label{section: Natural Modes}

To begin this analysis, we define the following terms. 
\(\mathbb{P}^{(l)} = E[\mathbf{X}^{(l)}\mathbb{D}^{(l)}]\) is the cross-correlation vector between the input and the desired response. \(\mathbb{E}^{(l)} = \mathbb{D}^{(l)} - \mathbb{Y}^{(l)}\) is the unrolled local error vector (Section \ref{section:Recast_CNN}). \(\overbar{\mathbb{R}}^{(l)} = E[(\mathbb{X}^{(l)}-\overbar{\mathbb{X}}^{(l)})\otimes (\mathbb{X}^{(l)}-\overbar{\mathbb{X}}^{(l)})]\) is the autocorrelation matrix. \(\overbar{\mathbb{X}}^{(l)}\) is the mean along the rows of \(\mathbf{X}^{(l)}\).\footnote{In the adaptive filter setting, we estimate the autocorrelation matrix by averaging the input vector's outer product with itself over many time steps, i.e., strides in time. In the restructured CNN, these strides are spatial. Therefore, we average over both the \(M\) columns of \(\mathbf{X}\) and a randomly sampled batch.} We use these terms to expand (\ref{eq:local_gradient_basic}):

\begin{equation} \label{eq:wiener_hopf}
    \nabla_{\mathbb{W}^{l}}J=E[\mathbf{X}^{(l)} \mathbb{E}^{(l)}]=\mathbb{P}^{(l)} - \mathbb{R}\mathbb{W}^{(l)} 
\end{equation}

Let us apply the principle of orthogonality to the CNN layer: when the estimation error vector, \(\mathbb{E}^{(l)}\), is orthogonal to the input \(\overrightarrow{\mathbf{X}}_{k,*}^{(l)}\) (and by extension, all rows of \(\mathbf{X}^{(l)}\)), then the weights are at their optimum value, \(\mathbb{W}^{(l)}_o\). At this point, the error vector is at its minimum, denoted as \(\mathbb{E}^{(l)}_o\), \(\nabla_l J = \nabla_{\mathbb{W}^{(l)}}J = 0\). In summary, applying the principle of orthogonality to (\ref{eq:local_gradient_basic}) results in \(E[\overrightarrow{\mathbf{X}}_{k,*}^{(l)} \mathbb{E}^{(l)}] = E[\mathbf{X}^{(l)} \mathbb{E}_o^{(l)}] = \mathbb{P}^{(l)} - \mathbb{R}^{(l)}\mathbb{W}_o^{(l)} = 0 \). This gives the Wiener-Hopf equations\(\mathbb{P}^{(l)} = \mathbb{R}^{(l)}\mathbb{W}_o^{(l)}\). We introduce the time step, \(n\), and \(\mu\) as the learning rate and apply the Wiener-Hopf equations and (\ref{eq:wiener_hopf}) to the weight update equation, \(\mathbb{W}^{(l)}(n+1) =\mathbb{W}^{(l)}(n) - \mu\nabla_{\mathbb{W}^{l}_k}J\): \(\mathbb{W}^{(l)}(n+1) = \mathbb{W}^{(l)}(n) + \mu(\mathbb{R}^{(l)}\mathbb{W}^{(l)}(n)- \mathbb{R}^{(l)}\mathbb{W}_o^{(l)}) \).

Define the weight error vector, \(\mathbb{C}^{(l)}(n)= \mathbb{W}^{(l)}(n) - \mathbb{W}^{(l)}_o\) and substitute into  this new weight update equation to obtain:
\begin{align} \label{weight_error_C_R}
    \mathbb{C}^{(l)}(n+1) = \mathbb{C}^{(l)}(n)[\mathbb{I} -\mu \mathbb{R}^{(l)}]
\end{align}

Define \(\mathbb{V}(n) = \mathbb{Q}^H\mathbb{C}(n)\). Substituting  the eigendecomposition \(\mathbb{R} = \mathbb{Q} \Lambda \mathbb{Q}^H\), where \(\Lambda \) is the diagonal matrix containing the eigenvalues of \(\mathbb{R}\), and \(\mathbb{V}(n)\) into (\ref{weight_error_C_R}) gives the transformed set of weight update equations: \(\mathbb{V}^{(l)}(n+1) = \mathbb{V}^{(l)}(n)[\mathbb{I} -\mu \Lambda^{(l)}]\). The natural modes are the elements in \(\mathbb{V}(n)\). Let \(v_k(n)\) be a single element in \(\mathbb{V}(n)\), indexed by \(k\), such that \(k =1,...,N\) for \(N\) eigenvalues and assume an initial starting point \(v_k^{(l)}(0)\). Then the natural modes are as follows:

\begin{equation}\label{eq:natural_modes_init}
    v_k^{(l)}(n) = (1-\mu \lambda_k)^n v^{(l)}_k(0)
\end{equation}

\subsection{Bounds of the Natural Modes}

The exponential form of (\ref{eq:natural_modes_init}) enables us to derive explicit bounds on stability and error decay rate. Training is stable when \(-1<1-\mu \lambda_k<1,\forall k\), indicating that any CNN architecture that uses a technique to suppress the largest eigenvalues is stable at comparatively higher \(\mu\). Assuming all modes have the same \(\mu\), the largest eigenvalue of all the modes sets the stability bound: 

\begin{equation}\label{eq:stability_bound}
    \mu < \frac{2}{\lambda_{max}}
\end{equation}

The time constant of the modes determine the speed of weight error decay, and by extension, the speed of training. Each entry in the weight error matrix \(\mathbb{C}(n)\) decays via an exponential \(e^{-\frac{n}{\tau_k}}\), where \(\tau_{k} = \frac{-1}{\ln(1-\mu \lambda_{k})}\). The bound on the training speed is set by the largest time constant, or the smallest eigenvalue:

\begin{equation}\label{eq:speed_bound}
    \tau_{max} = \frac{-1}{\ln(1-\mu \lambda_{min})}
\end{equation}

Equation (\ref{eq:speed_bound}) shows that training converges faster when using techniques that boost the smallest eigenvalues without compromising stability. 

Increased stability and training speed are well-established benefits of whitening. True matrix whitening is not typically performed in both the traditional adaptive filter domain and the forward path of CNNs, because the whitening matrix calculation is cost-prohibitive. BatchNorm improves stability and increases training speed because BatchNorm is a cost-effective approximation of whitening. 

\subsection{CNN Spatial Strides and Intra-Channel Statistics} \label{section: Channel_strides_statistics}

\begin{figure}[t]
\centering

\begin{subfigure}[t]{0.18\textwidth}
  \centering
  \includegraphics[width=1\linewidth]{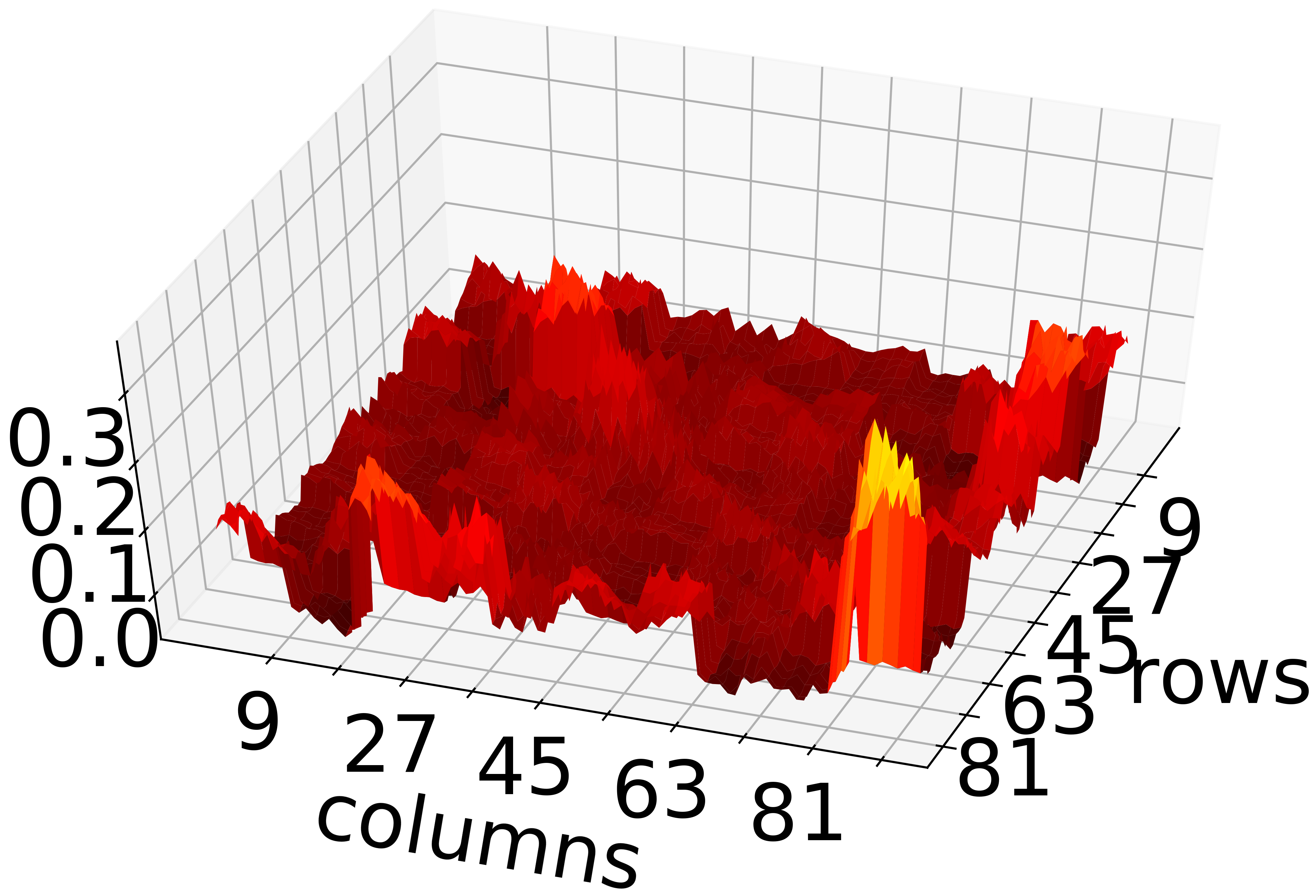}
  \subcaption{Without BatchNorm}
  \label{fig:3D_plot_no_batchnorm}
\end{subfigure}
\begin{subfigure}[t]{0.18\textwidth}
  \centering
  \includegraphics[width=1\linewidth]{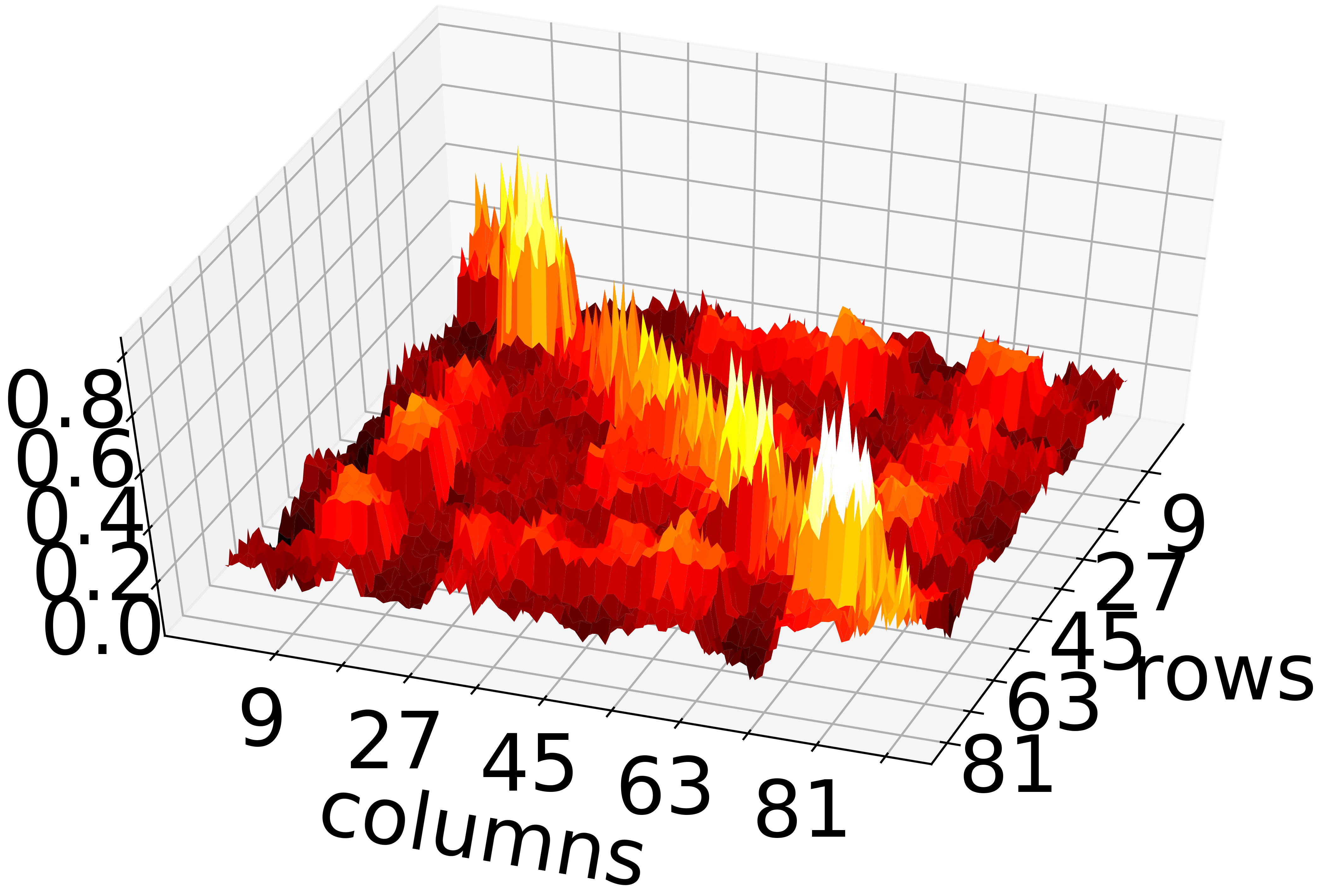}
  \subcaption{With BatchNorm}
  \label{fig:3D_plot_batchnorm}
\end{subfigure}
\caption{Heatmap of the autocorrelation of the first 90 elements in the unrolled input vector of second residual block of ResNet18.}
\label{fig:AutoCorr_Block2}
\end{figure}

\begin{figure}[t]
  \centering
  \includegraphics[width=0.48\linewidth]{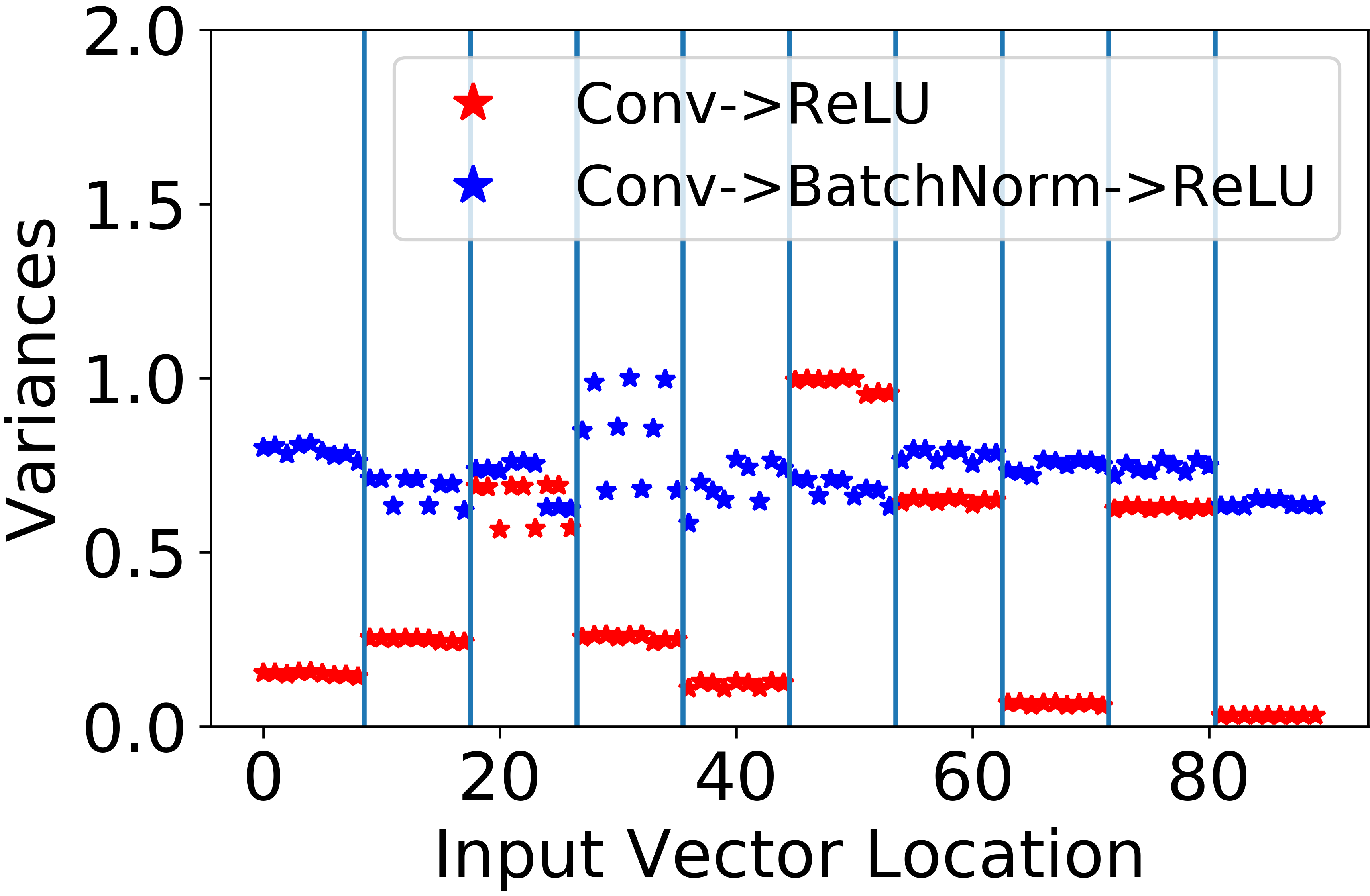}
\caption{Variances for the first 90 elements of the unrolled input vector of the first residual block. Vertical lines separate elements from different channels. ``Conv\(\rightarrow\)ReLU'' is from a ResNet18 without BatchNorm. ``Conv\(\rightarrow\)BatchNorm\(\rightarrow\)ReLU'' is from a ResNet18 with BatchNorm. Batch size is 256.}
\label{fig:ChannelVariances_Comparison}
\end{figure}

\begin{figure}[t]
\centering
\begin{subfigure}[t]{0.18\textwidth}
  \centering
  \includegraphics[width=0.55\linewidth]{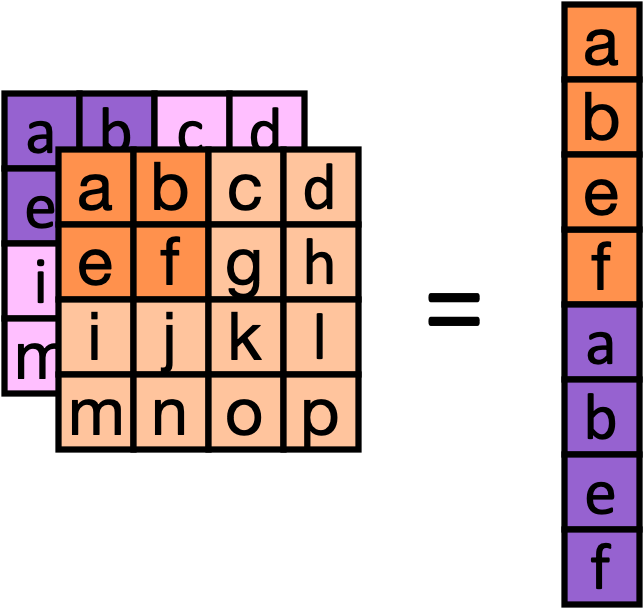}
  \subcaption{Before the first stride}
  \label{fig:unrolled_input_1}
\end{subfigure}\hfil
\begin{subfigure}[t]{0.18\textwidth}
  \centering
  \includegraphics[width=0.55\linewidth]{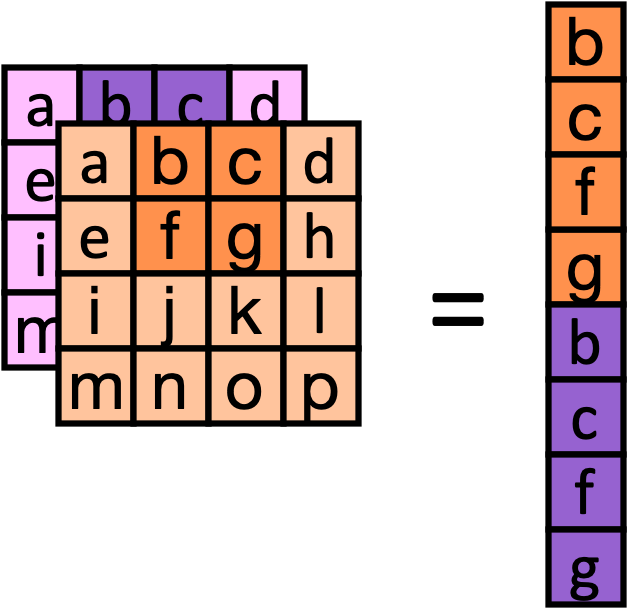}
  \subcaption{After the first stride}
  \label{fig:unrolled_input_2}
\end{subfigure}
\caption{Unrolled 3D feature map patches. Pixels from the same channel remain within the boundaries of their respective block rows.}
\label{fig:unrolled_input_multi_channel}
\end{figure}

BatchNorm is a pseudo-whitening operation because it partially decorrelates the input without an explicit calculation of the whitening matrix. Fig. \ref{fig:AutoCorr_Block2} illustrates this effect. A realistic set of input activations (not statistically white) is generated from ResNet18 \cite{He2016} trained on CIFAR-10 \cite{krizhevsky2009learning} without BatchNorm and shown in Fig. \ref{fig:3D_plot_no_batchnorm}. These activations have strong off-diagonal values and weak main diagonal values and show a block pattern. Each block's side-length is equal to the number of vector elements corresponding to a single channel. This pattern emerges from the uniformity of the intra-channel statistics and wide variation in inter-channel statistics (Fig. \ref{fig:ChannelVariances_Comparison}). The row-wise blocking (illustrated in Fig. \ref{fig:unrolled_input_multi_channel}) and overlapping strides result in elements of the same block row having similar statistics. BatchNorm, at the granularity of blocks, transforms the activations to resemble a whitened input by strengthening the blocks along the main diagonal while weakening the off-diagonal blocks (Fig. \ref{fig:3D_plot_batchnorm}). 

The matrix \(\mathbb{R}\) is the concatenation of block matrices, where each block matrix is the autocorrelation matrix of block rows. In the extreme case where there is no correlation between channels, each block is responsible for a different set of eigenvalues. Entire groups of eigenvalues can be scaled by scaling their associated channel.

\begin{figure}[tbp]
\centering
\begin{subfigure}[ht]{0.15\textwidth}
  \centering
  \includegraphics[width=1\linewidth]{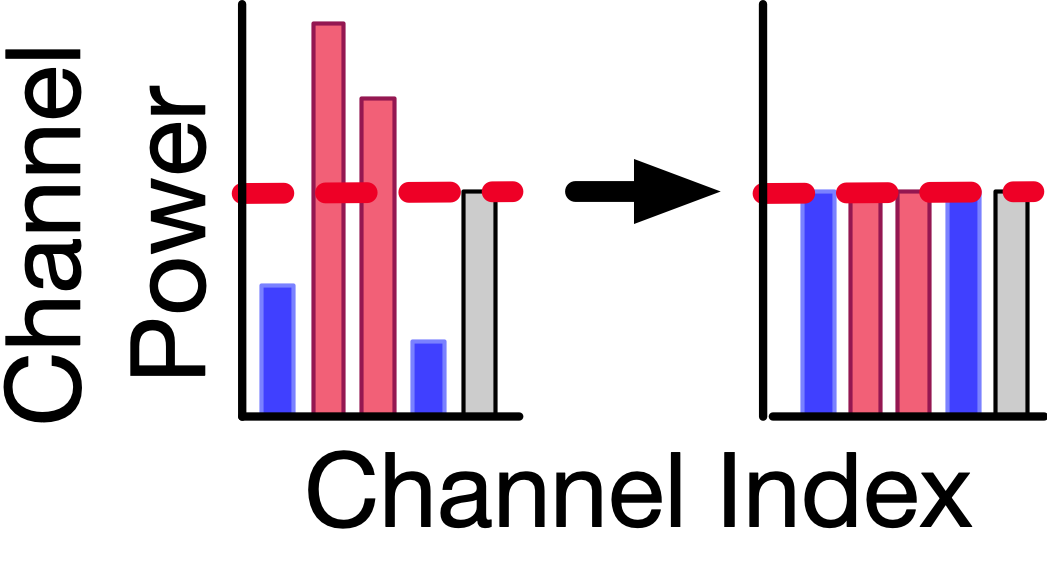}
  \caption{BatchNorm}
  \label{fig:Cartoon_BatchNorm}
\end{subfigure}
\begin{subfigure}[ht]{0.15\textwidth}
  \centering
  \includegraphics[width=1\linewidth]{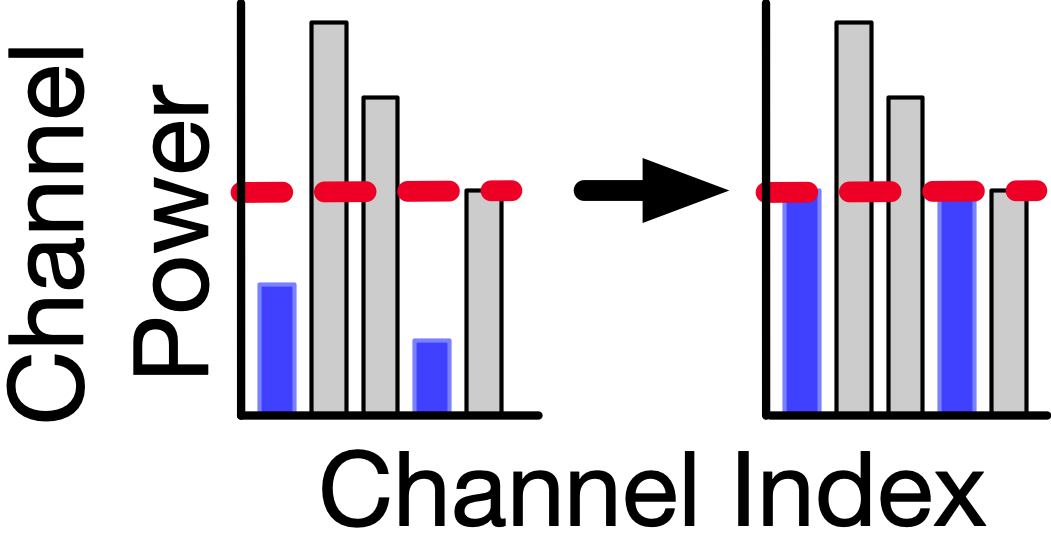}
  \caption{BN\_Amplify}
  \label{fig:Cartoon_BN_Amplify}
\end{subfigure}
\begin{subfigure}[ht]{0.15\textwidth}
  \centering
  \includegraphics[width=1\linewidth]{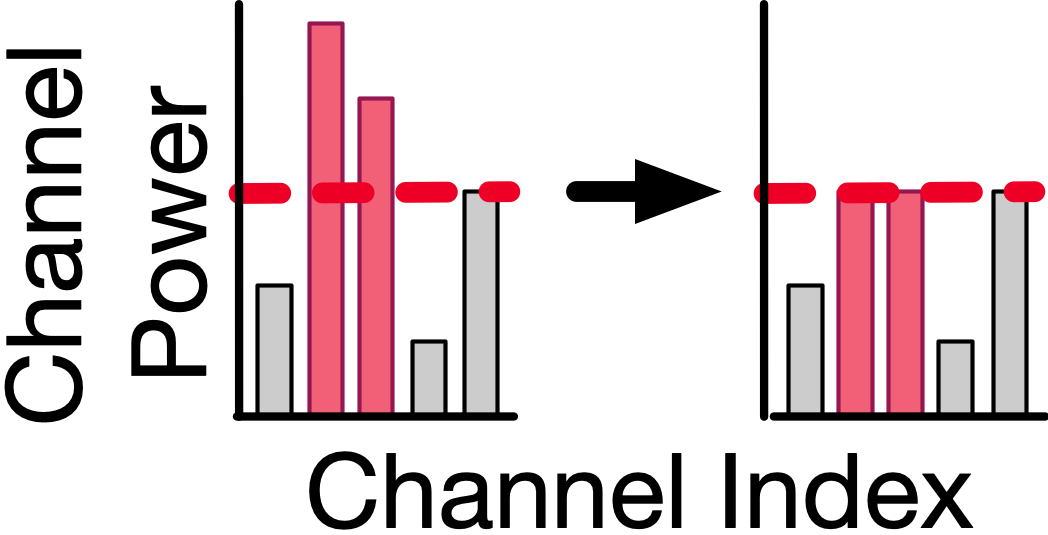}
  \caption{BN\_Suppress}
  \label{fig:Cartoon_BN_Suppress}
\end{subfigure}
\caption{Proposed variants of BatchNorm: BN\_Amplify and BN\_Suppress. Dashed line = power threshold, pink = above threshold, and purple = below threshold.}
\label{fig:cartoon_BN_Variants}
\end{figure}

While in reality, there is some correlation between channels, large eigenvalues primarily result from the channels with larger power, and small eigenvalues result from the channels with smaller power. Therefore, there exists the possibility that we can independently tune the large and small eigenvalues by normalizing only the largest and smallest channels, respectively. This ability allows us to delineate the effects of the bounds proposed in Section \ref{section: Natural Modes}. Therefore, we propose two BatchNorm variants: BN\_Amplify and BN\_Suppress. BN\_Amplify amplifies channels using normalization only on channels above a power threshold of 1.0. BN\_Suppress suppresses channels using normalization only on channels below a power threshold of 1.0 (Fig. \ref{fig:cartoon_BN_Variants}).

\section{Experiments}\label{section: Experiment_NaturalModes}
\subsection{MNIST}\label{section:MNIST_Modes}

The networks are based on LeNet \cite{le1989handwritten} with one FC layer removed (to reduce memory requirements for saving activations from multiple training steps). We compare four networks:

\begin{itemize}
    \item Baseline: Described in Table \ref{Tab:Baseline}
    \item BatchNorm: BatchNorm layer after each conv layer
    \item BN\_Amplify: BN\_Amplify layer after each conv layer
    \item BN\_Suppress: BN\_Suppress layer after each conv layer
\end{itemize}

All networks are trained with five seeds, starting with the same set of random weights on the MNIST dataset \cite{lecun1998gradientMNIST}. \(\mu\) for the convolution weights is swept over 20 epochs. We capture the minimum and maximum eigenvalues after the five steps across various \(\mu\). All other parameters are trained at \(\mu=0.1\) without dropout. The training algorithm uses stochastic gradient descent and cross-entropy loss.

\begin{table}[tbp]
 \captionof{table}{Baseline architecture\label{Tab:Baseline}}
\centering
 \begin{tabular}{c c c c} 
 \hline
Layer & Type & Filter Size & Stride\\
 \hline
 1 & Conv & 1\(\times\)6\(\times\)5\(\times\)5 & 1 \\
 2 & ReLU & - & - \\
 3 & Avg Pool & 2\(\times\)2 & 2 \\
 4 & Conv & 6\(\times\)16\(\times\)5\(\times\)5 & 1 \\
 5 & ReLU & - & - \\
 6 & Avg Pool & 2\(\times\)2 & 2 \\
 7 & Fully Connected & 400\(\times\)10 & - \\
 \hline
 \end{tabular}
\end{table}

\begin{figure}[tbp]
    \centering
    \includegraphics[width=0.48\linewidth]{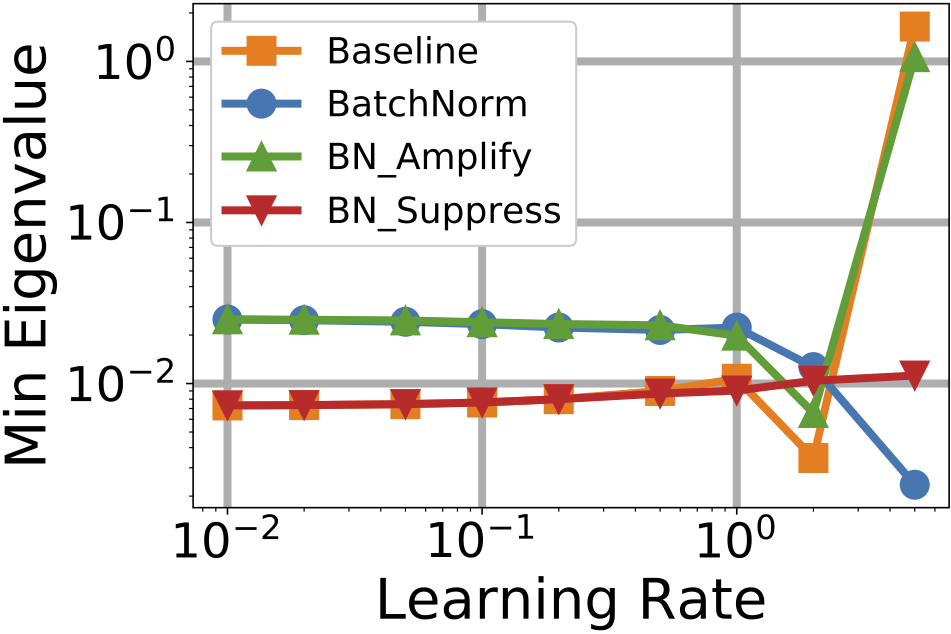}
    \caption{Minimum eigenvalues of \(\mathbb{R}\) after 5 training steps.}
    \label{fig:Min_Eigenvalues}
\end{figure}

\begin{figure}[b]
\centering
\begin{subfigure}{0.2\textwidth}
\captionsetup{width=0.8\textwidth}
  \centering
  \includegraphics[width=1\linewidth]{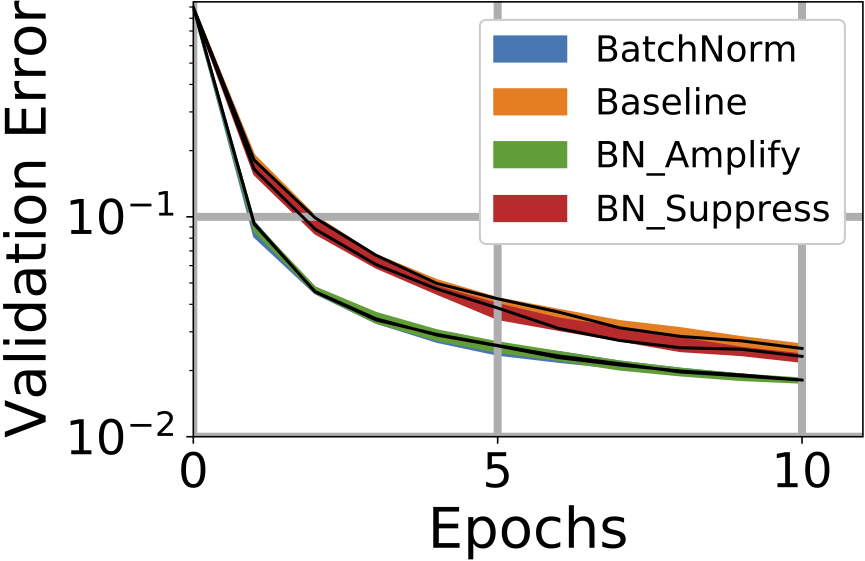}
  \caption{\(\mu\) = 0.01}
  \label{fig:Convergence Speed Training Curves 0.01}
\end{subfigure}
\begin{subfigure}{0.2\textwidth}
\captionsetup{width=0.8\textwidth}
  \centering
  \includegraphics[width=1\linewidth]{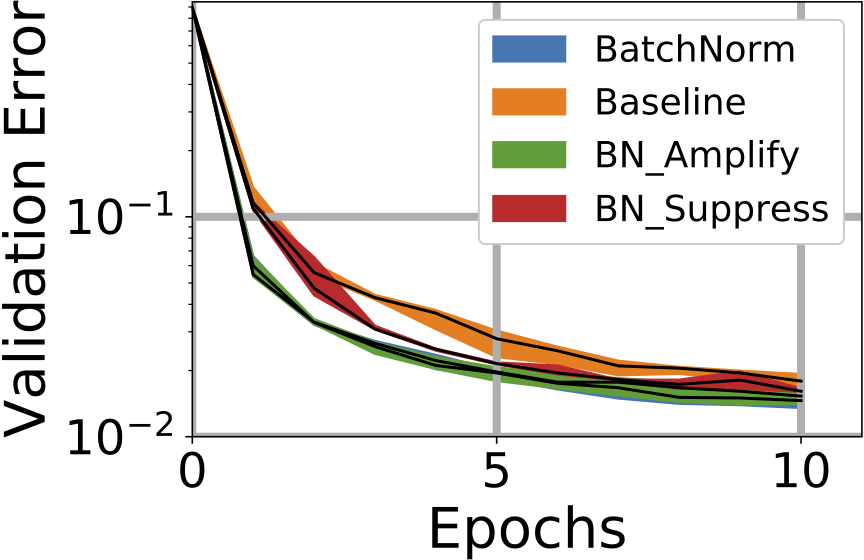}
  \caption{\(\mu\) = 0.1}
  \label{fig:Convergence Speed Training Curves 0.1}
\end{subfigure}\\
\begin{subfigure}{0.2\textwidth}
\captionsetup{width=0.8\textwidth}
  \centering
  \includegraphics[width=1\linewidth]{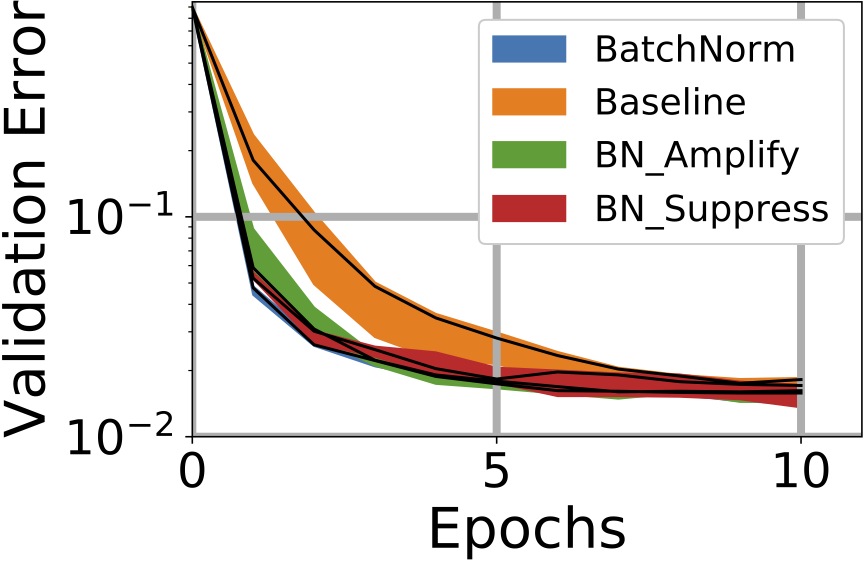}
  \caption{\(\mu\) = 0.5}
  \label{fig:Convergence Speed Training Curves 0.5}
\end{subfigure}
\begin{subfigure}{0.2\textwidth}
\captionsetup{width=0.8\textwidth}
  \centering
  \includegraphics[width=1\linewidth]{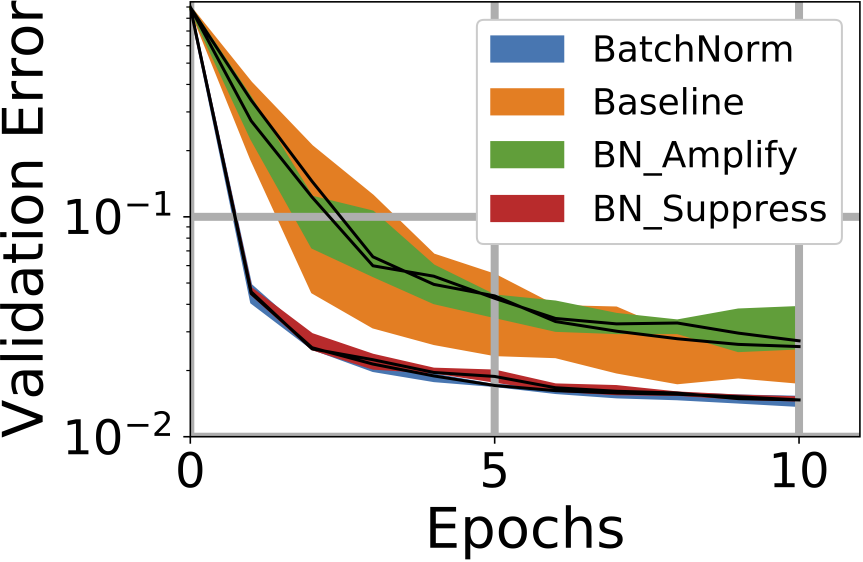}
  \caption{\(\mu\) = 1.0}
  \label{fig:Convergence Speed Training Curves 1.0}
\end{subfigure}\\
\caption{Training curves at different \(\mu\). Color bands denote \nth{2} to \nth{3} quartile spread of validation error from five seeds.}
\label{fig:Convergence Speed Training Curves}
\end{figure}

\begin{figure}[b]
\centering
\begin{subfigure}{0.25\textwidth}
\captionsetup{width=0.9\textwidth}
    \centering
    \includegraphics[width=1\linewidth]{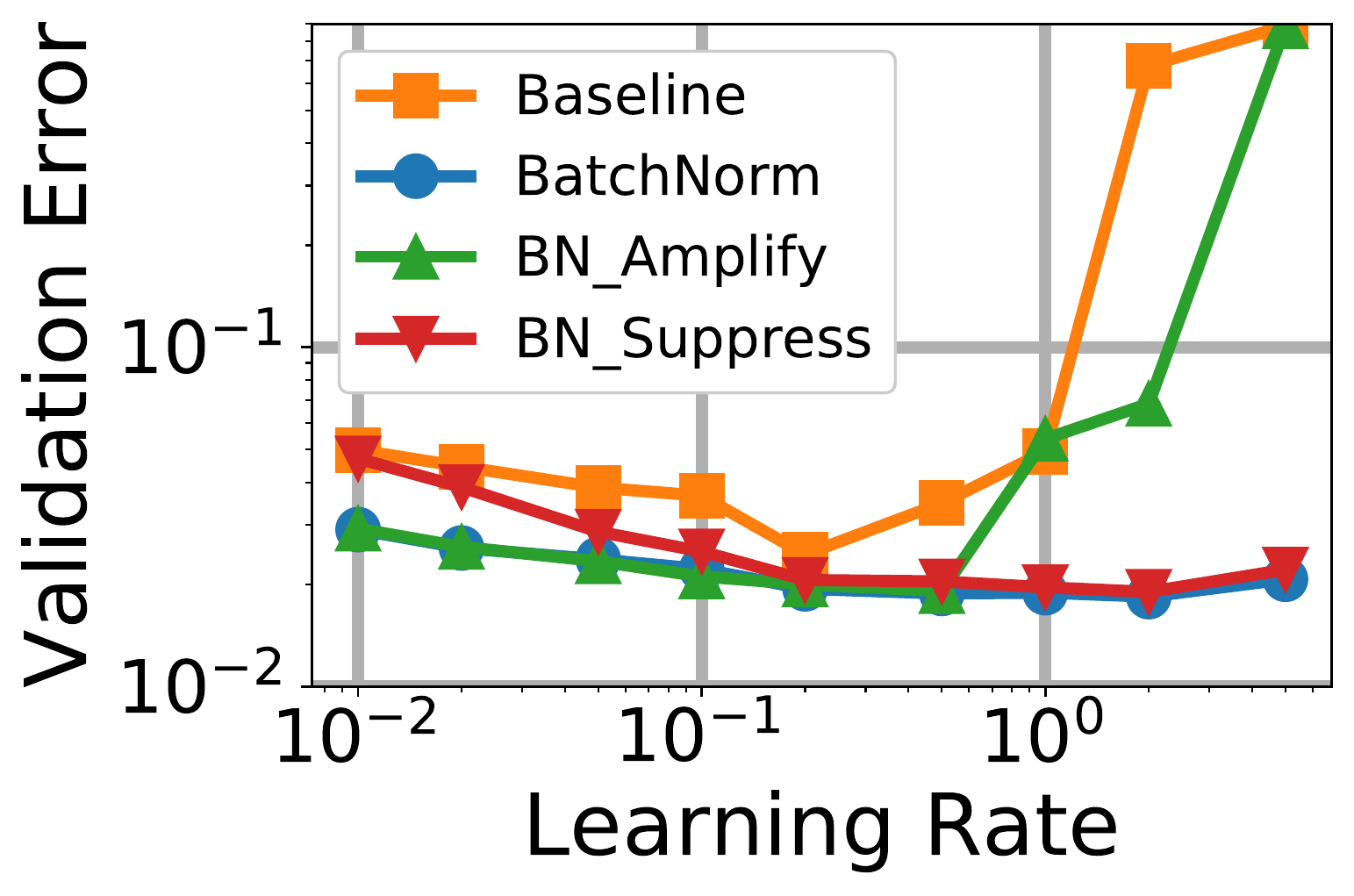}
    \caption{Validation error vs. \(\mu\) (20 epochs, 5 seeds)}
    \label{fig:Training Curve Stability}
\end{subfigure}%
\begin{subfigure}{0.23\textwidth}
\captionsetup{width=0.9\textwidth}
    \centering
    \includegraphics[width=1\linewidth]{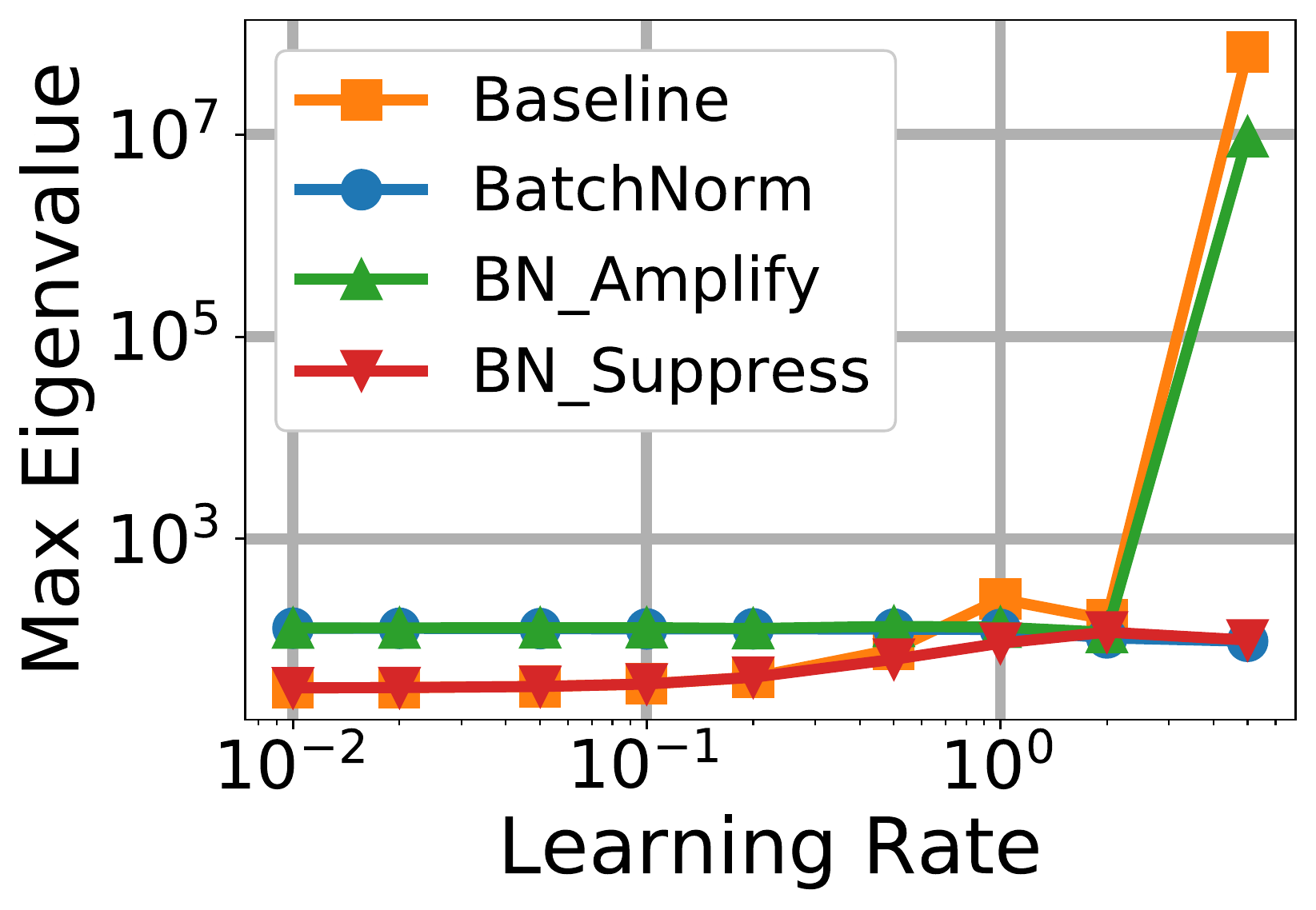}
    \caption{Max eigenvalues of \(\mathbb{R}\) vs. \(\mu\) (after 5 steps)}
    \label{fig:Max_Eigenvalues}
\end{subfigure}
\caption{Stability experiments.}
\label{fig:Stability Curves}
\end{figure}

\subsubsection{Results for Training Convergence Speed Measurements on MNIST}
The nearly indistinguishable BatchNorm and BN\_Amplify curves in Fig. \ref{fig:Convergence Speed Training Curves 0.01} and Fig. \ref{fig:Convergence Speed Training Curves 0.1} show that at low \(\mu\), BatchNorm is amplifying the weaker channels. In Fig. \ref{fig:Min_Eigenvalues} below \(\mu=0.1\), the minimum eigenvalues of BatchNorm and BN\_Amplify not only match, but they are higher than the minimum eigenvalues of the other curves. 

Amplification did not affect the final validation error for \(\mu<0.5\). Even though we plot up to 20 epochs, all networks eventually reached the same validation error in all cases for \(\mu<0.5\). This observation reinforces the observation that at these lower \(\mu\), the amplification of the smallest eigenvalues benefits primarily speed rather than absolute validation error.

\subsubsection{Stability Analysis on MNIST}

At high \(\mu\), the primary benefit of BatchNorm is in suppressing the largest eigenvalue. In Fig. \ref{fig:Convergence Speed Training Curves 1.0}, the validation errors of BatchNorm and BN\_Suppress are nearly indistinguishable, indicating at these high \(\mu\), the primary benefit of BatchNorm is in suppression. Indeed, in Fig. \ref{fig:Max_Eigenvalues} at \(\mu < 1.0\), the maximum eigenvalue of BatchNorm and BN\_Suppress is nearly indistinguishable from that of BN\_Suppress. In contrast, BN\_Amplify and Baseline cannot to suppress the largest eigenvalues and therefore become unstable.

Figs. \ref{fig:Convergence Speed Training Curves} and \ref{fig:Training Curve Stability} show that there is a crossover point between \(\mu=0.1\) and \(\mu=1.0\). In this range, the stability bound begins to dominate training performance over the slowest mode bound, and for BN\_Amplify and Baseline, the validation error starts increasing.  This growing instability is marked in Fig. \ref{fig:Convergence Speed Training Curves 0.5} and \ref{fig:Convergence Speed Training Curves 1.0} by an increase in the error bands. 

The toy network described in Table \ref{Tab:Baseline} provides a controlled environment to disentangle the dynamics of BatchNorm and its effects on the bounds of the natural modes. However, due to the strong dependence modern neural networks have on BatchNorm, we expand this analysis to ResNet20.

\subsection{ResNet20 on CIFAR-10}

ResNet20 \cite{He2016}, featuring residual connections that allows for much deeper networks, is difficult to train without BatchNorm. To stably train the BatchNorm variants of interest, we use the core portion of Fix-Up initialization \cite{Zhang2019} which applies depth-wise layer scaling of the weights to our ResNet models. We call this "FixupCore" in our experiments. This scaling removes the exponential scaling effect of depth on the variances and allows the more detailed study of the impact of channel-wise scaling of BatchNorm on the bounds provided by the Natural Modes analysis (Fig. \ref{fig:variance_ResNet20}). We compare five networks:

\begin{itemize}
    \item FixupCore: ResNet20 with FixupCore
    \item BN\_Suppress: ResNet20 with BN\_Suppress
    \item BN\_Suppress\_FixupCore: ResNet20 with BN\_Suppress, FixupCore initialization
    \item BatchNorm: ResNet20 with BatchNorm
    \item BN\_Amplify\_FixupCore: ResNet20 with BN\_Amplify, FixupCore initialization
\end{itemize}

All networks are trained with 20 seeds for 20 epochs without dropout. We sweep \(\mu\) for the convolution weights, but fix \(\mu=0.1\) for all other parameters. 

\subsubsection{Training Without BatchNorm: The Connection to Initialization}

Prior to BatchNorm, CNNs were deemed very sensitive to initialization. Networks such as AlexNet \cite{Krizhevsky2012AlexNet} and VGG \cite{simonyan2014very} were developed before the introduction of BatchNorm but were sensitive to the selection of \(\mu\) and weight initialization. Works studying neural network initialization \cite{glorot2010understanding}\cite{he2015delving} aim to reduce this sensitivity by using the heuristics of the layer-wise variance of activations and gradients to set the weights during initialization. The authors hoped to strike a balance between the problems of exploding and vanishing gradients. BatchNorm reduces CNNs' reliance on these methods.

\begin{figure}[t]

\end{figure}

\begin{figure}[t]
\begin{subfigure}[t]{0.2\textwidth}
  \centering
  \includegraphics[width=1\linewidth]{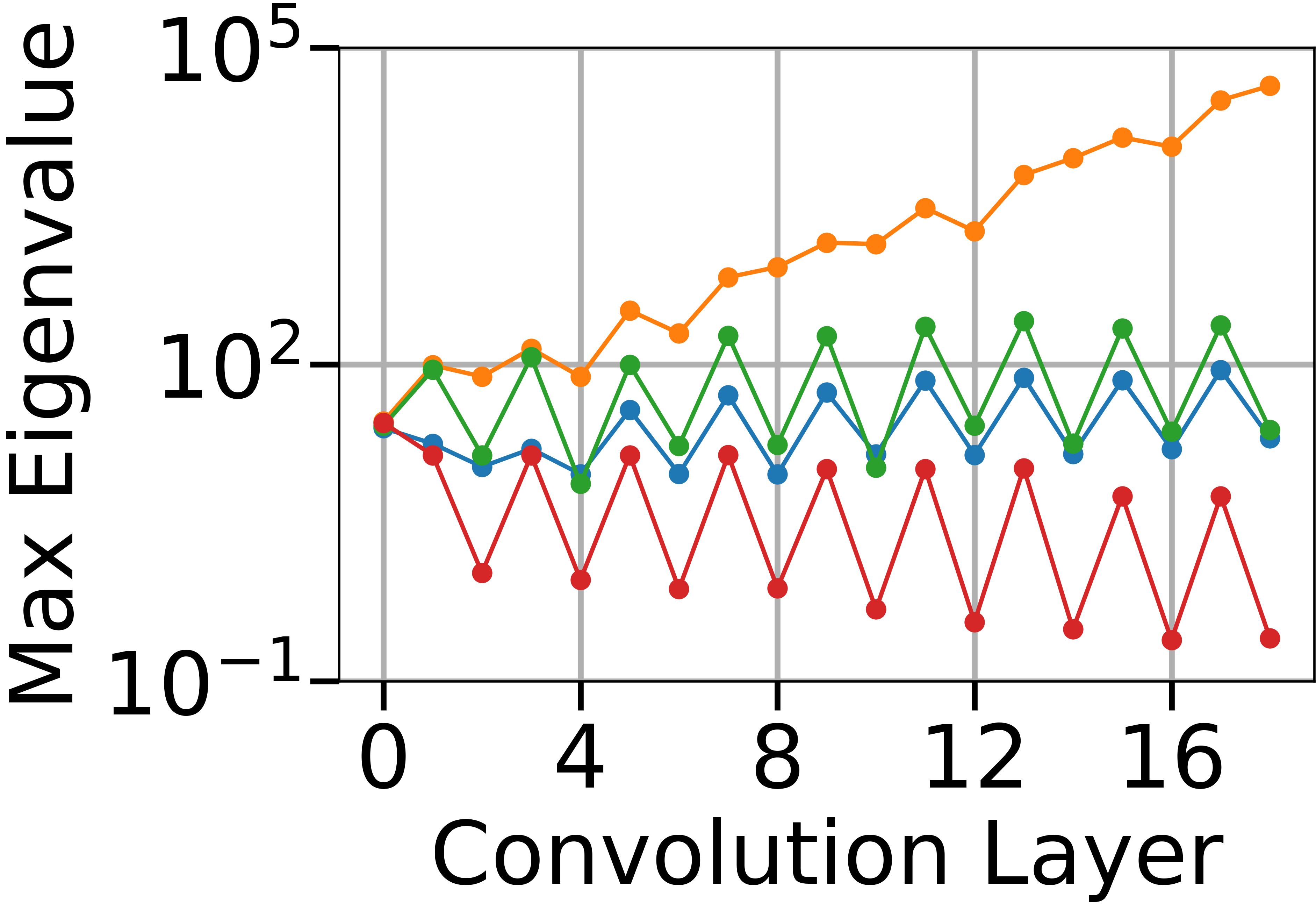}
  \subcaption{Maximum eigenvalues}
  \label{fig:first_step_MaxEig_ResNet20}
\end{subfigure}\hfil
\begin{subfigure}[t]{0.2\textwidth}
\centering
  \includegraphics[width=1\linewidth]{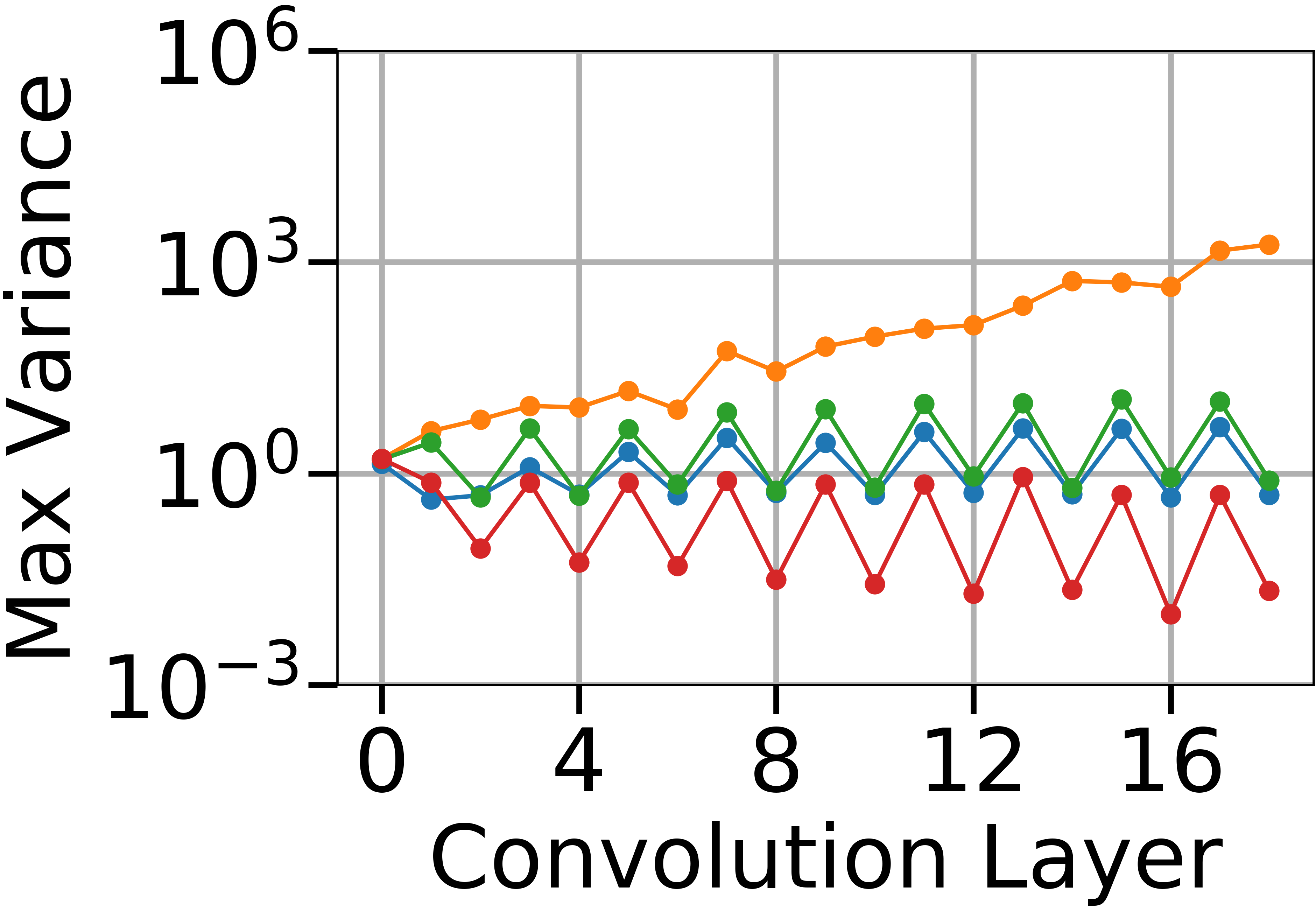}
\caption{Maximum variance}
\label{fig:variance_ResNet20}
\end{subfigure}\\
\begin{subfigure}[t]{0.5\textwidth}
  \centering
  \includegraphics[width=1\linewidth]{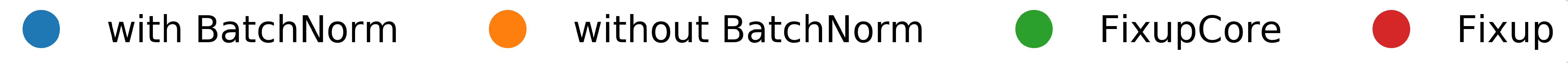}
\end{subfigure}
\caption{Maximum variance of the input vector elements and the maximum eigenvalues for the autocorrelation matrix of the convolution layers for the first forward pass of ResNet20. Batch size is 128.}
\label{fig:first_step_Eigenvalues_ResNet20}
\end{figure}

Natural mode analysis presents a more concrete interpretation of the initialization story. The analysis shows that the stability bound is set by the largest eigenvalue and \(\mu\). Since the common method for training neural networks is to set all layers of the network to have the same \(\mu\), the maximum allowed eigenvalue must also be the same for all layers of the network. Limiting the variance is a very coarse method to limit the maximum eigenvalue because the maximum eigenvalue is upper-bounded by the variance. Indeed, in Fig. \ref{fig:first_step_MaxEig_ResNet20}, the maximum eigenvalue exponentially increases with depth without Fixup or normalization.

\subsubsection{Training Speed Experiments}

\begin{figure}[b]
\centering
\begin{subfigure}[t]{0.2\textwidth}
\captionsetup{width=1.0\textwidth}
  \centering
  \includegraphics[width=1\linewidth]{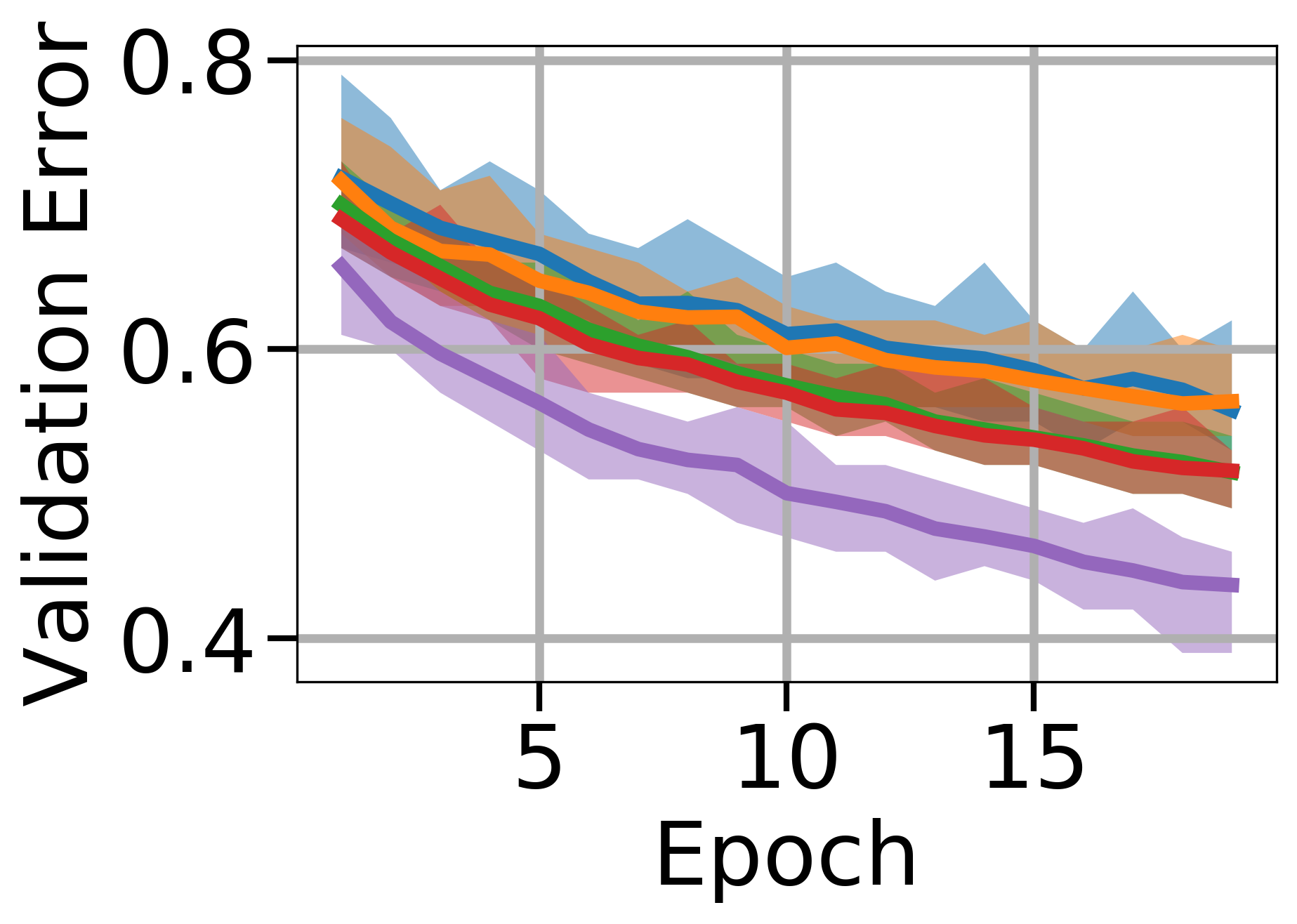}
  \caption{\(\mu\) = 0.0001}
  \label{fig:Resnet20 Convergence Speed Training Curves 0.005}
\end{subfigure}
\begin{subfigure}[t]{0.21\textwidth}
\captionsetup{width=1.0\textwidth}
\centering
  \includegraphics[width=1\linewidth]{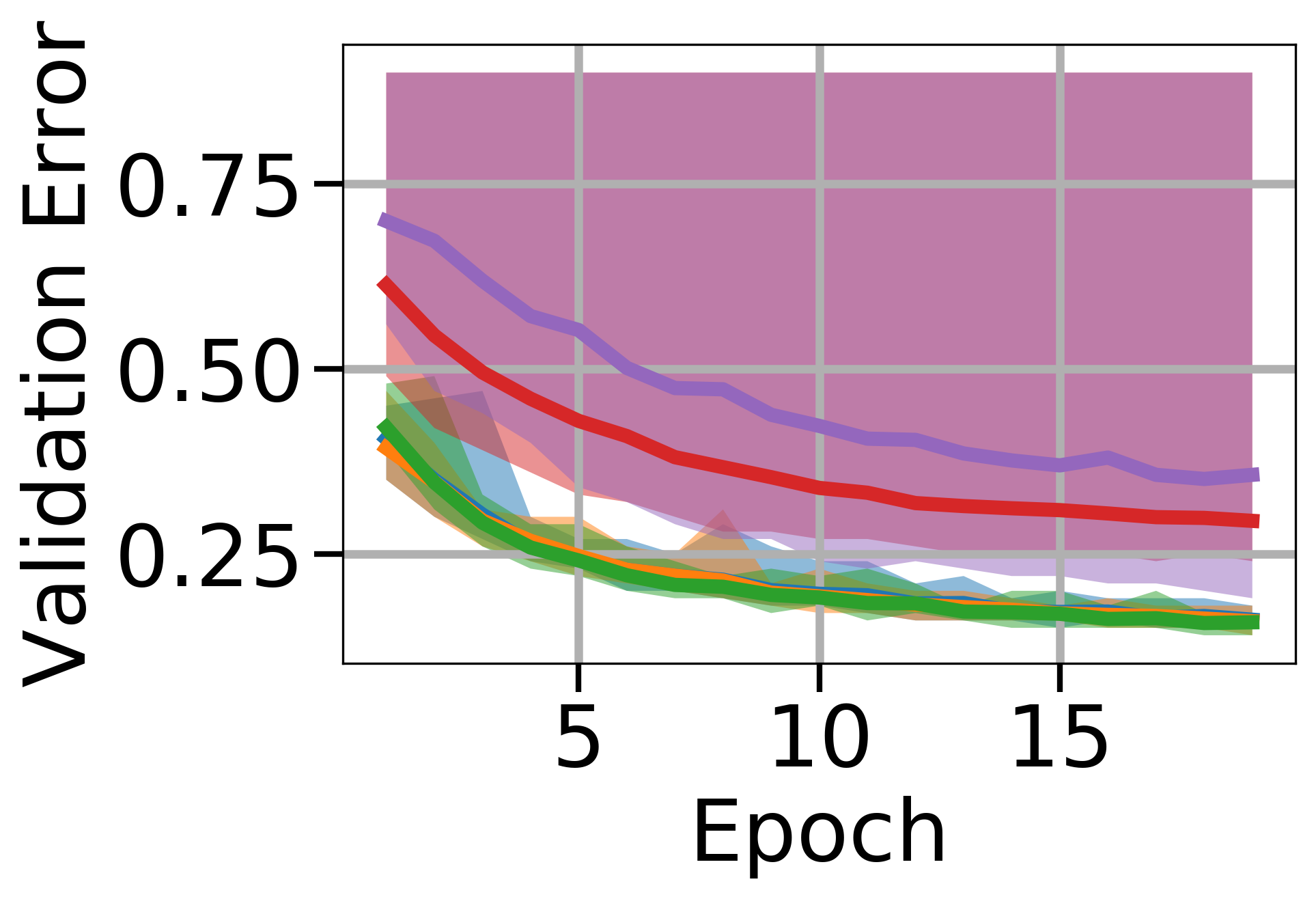}
  \caption{\(\mu\) = 0.05}
  \label{fig::Resnet20 Convergence Speed Training Curves 0.05}
\end{subfigure}
\begin{subfigure}[t]{0.5\textwidth}
\centering
  \includegraphics[width=1\linewidth]{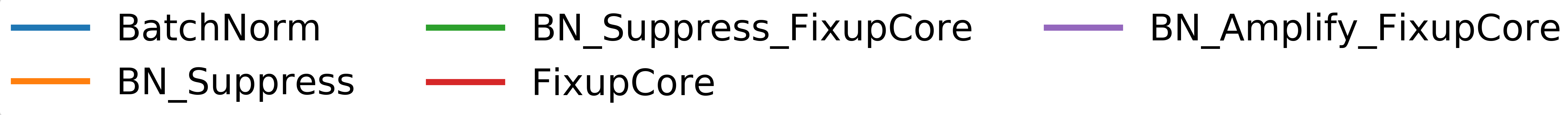}
\end{subfigure}
\caption{Training curves at different \(\mu\) for Resnet20 on CIFAR-10. Bands denote \nth{2} to \nth{3} quartile spread of the error.}
\label{fig::Resnet20 Convergence Speed Training Curves}
\end{figure}

We plot the training curves at different \(\mu\) (Fig. \ref{fig::Resnet20 Convergence Speed Training Curves}). For low \(\mu\), BN\_Amplify\_FixupCore trains the fastest. However, since BatchNorm could not match the performance of BN\_Amplify, we cannot conclude that for residual networks that BatchNorm’s primary function at low \(\mu\) is to amplify weak channels. Further work is needed to understand BatchNorm’s effect on the minimum eigenvalues at these low learning rates.

\subsubsection{Stability}

\begin{figure}[tbp]
\centering
\begin{subfigure}[t]{0.2\textwidth}
\captionsetup{width=0.9\textwidth}
    \centering
    \includegraphics[width=1\linewidth]{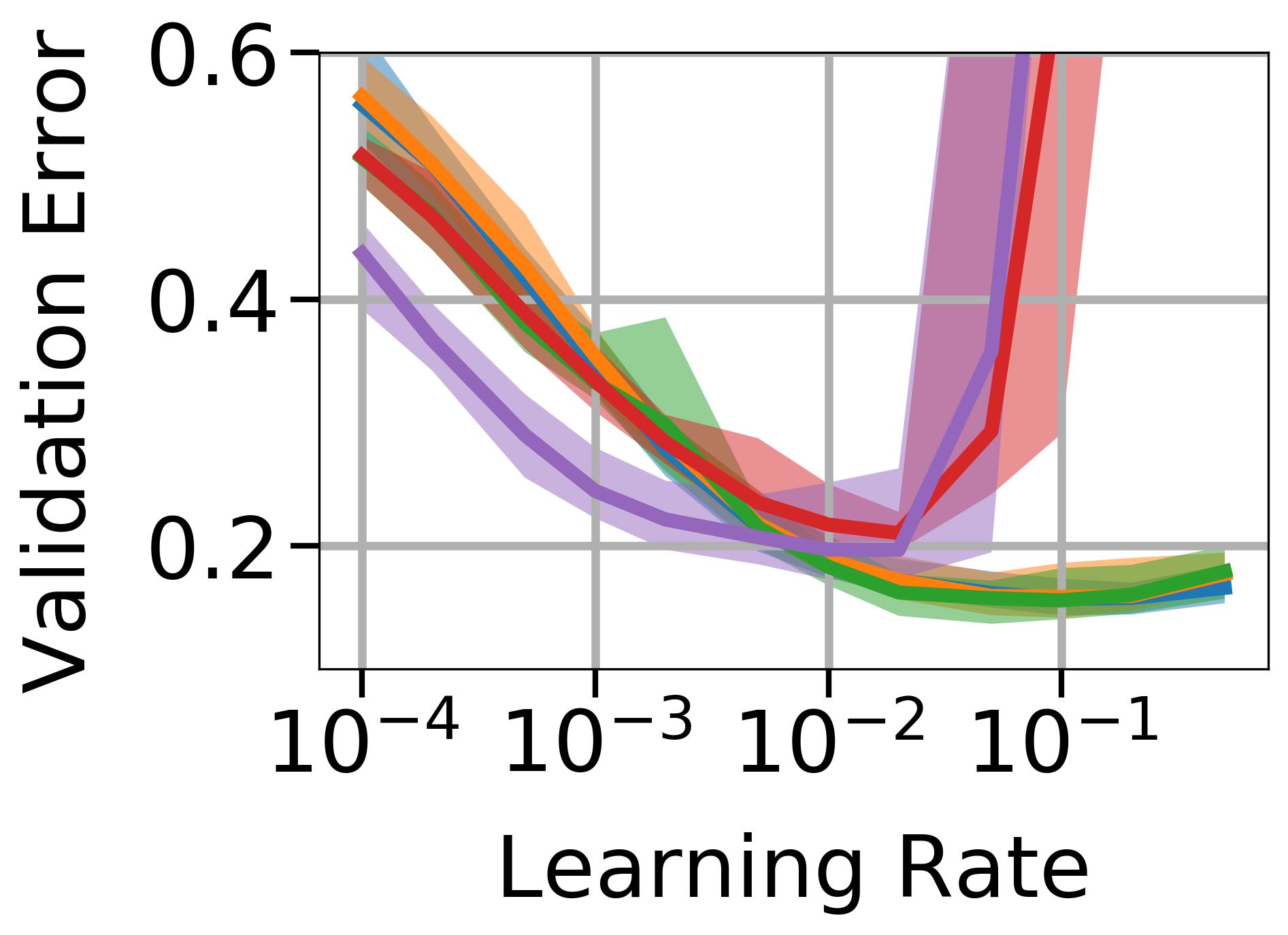}
    \caption{Validation error vs. \(\mu\) (19 epochs, 20 seeds). Bands denote \nth{2} to \nth{3} quartile spread of the error.}
    \label{fig:ResNet20_Stability_val_error}
\end{subfigure}\hfil
\begin{subfigure}[t]{0.2\textwidth}
\captionsetup{width=0.9\textwidth}
    \centering
    \includegraphics[width=1\linewidth]{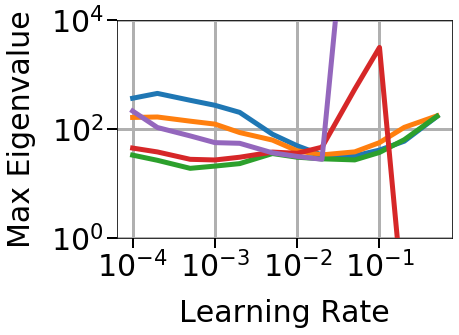}
    \caption{Max eigenvalues of \(\mathbb{R}\) vs. \(\mu\) (after 19 epochs) for inputs for 3rd residual block.}
    \label{fig:ResNet20_Stability_MaxEig}
\end{subfigure}
\begin{subfigure}[t]{0.5\textwidth}
\centering
  \includegraphics[width=1\linewidth]{Images/BatchNorm_ResNet20_Epochs/legend_3.png}
\end{subfigure}%
\caption{Stability experiments for ResNet20 on CIFAR-10.}
\label{fig:Resnet20_Stability Curves}
\end{figure}

We explore the stability of the network at different \(\mu\) (Fig. \ref{fig:Resnet20_Stability Curves}). At \(\mu=0.05\), the two networks without active suppression via normalization become unstable. Fixup does not help because it is an initialization technique. Fig. \ref{fig:Resnet20_Stability Curves} shows how the eigenvalues rapidly increase before the network fails to train completely.

\section{NLMS and the Principle of Minimum Disturbance} \label{section: NLMS}

This section details the similarities and differences between NLMS and BatchNorm. We formally introduce NLMS as a result of an optimization problem based on the Principle of Minimum Disturbance (PMD) \cite{Haykin:2002}. We demonstrate how to apply PMD to CNNs, and that BatchNorm placed before the convolution operation most closely matches NLMS. We conclude with noise injection experiments to validate the theoretical results.

LMS is a popular stochastic gradient algorithm used to find the filter weights of an adaptive filter based on just the current input and current error \cite{widrow1960adaptive}: \(\mathbb{W}(n+1) = \mathbb{W}(n) + \varepsilon(n) \mathbb{X}(n)\). A well-known problem with LMS is its sensitivity to the input size, where large inputs amplify gradient noise. As a result of this sensitivity, when the input power fluctuates, it becomes hard to maintain stability at high \(\mu\). Normalized LMS (NLMS) is a technique that resolves this sensitivity by dividing the weight update by the input power: \(\mathbb{W}(n+1) = \mathbb{W}(n) + \frac{\varepsilon(n)}{||\mathbb{X}(n)||^2} \mathbb{X}(n)\).

NLMS is derived using the Principle of Minimum Disturbance. PMD is the idea that while there are many ways to update the weights to reach the desired output, the best collection of updates is the one that disturbs the weights the least. For a traditional adaptive filter with a scalar output and vector weights and inputs, one should minimize \(J_{PMD}(n+1) = ||\delta_\mathbb{W}(n+1)||^2 = \sum_{k=0}^{K-1}(w_k(n+1)-w_k(n))^2\) subject to the constraint \(d(n) = \mathbb{W}^T(n+1)\mathbb{X}(n) = \sum_{k=0}^{K-1}(w_k(n+1)x_k(n))\).

We now prove that PMD can be applied to the CNN weight update. Since we cannot derive \(\mathbb{D}^{(l)}\) directly from the global CNN loss function \(J_\theta\), we derive \(\mathbb{D}^{(l)}\) from the local error provided by backpropagation. Because we are constraining our analysis to the components and operations in a single layer, there is no need to explicitly index by layer. Therefore, we drop references to the layer index \(l\).

\subsubsection{The Scalar Output Case}
Constrain the analysis to a single pixel in a given output feature map, indexed by \(m\). Then the output is scalar, and the weight update can be expressed using ordinary LMS. The output pixel at location \(m\) is given by \(y^{(m)}(n)\) at time step \(n\). 

To apply the PMD to the convolution layer, we need the desired output \(d^{(m)}(n)\). For a traditional adaptive filter, the desired response is provided externally by the system that the filter is trying to model. In the convolution layer case, internal layers do not have a local desired response directly provided. Instead, they have the local error \(\delta^{(m)}(n)\). Therefore, we assume the following relationship to derive the desired response: \(d^{(m)}(n) =  y^{(m)}(n) -\delta^{(m)}(n)\). Using \(d^{(m)}(n)\), NLMS can be derived for this scalar case.

\subsubsection{Multiple Input Channels}
Recall that for every output pixel \(y^{(m)}(n)\), where \(m \in [0,...,(OH\texttt{x}OW)-1]\), there is an unrolled input patch \(\mathbb{X}^{(m)}(n)\), with size \(K = IC\texttt{x}H\texttt{x}W\). However, there is a distinct set of \(Z = H\texttt{x}W\) weights for the block row of weights corresponding to channel \(i\). Using the components defined in Section \ref{section:Recast_CNN}, and notation developed in Section \ref{section: Natural Modes} for block rows in \(\mathbb{X}\), we can index into this patch and the associated weight element with \(z\), where \(z \in [0,...,Z-1]\) for input channel \(i\):

\begin{align} \label{eq:PMD_CNN_scalar}
    w_{z,i}(n+1) & = w_{z,i}(n) + \frac{\delta^{(m)}(n)}{||\mathbb{X}_i^{(m)}(n)||^2} x_{k,i}^{(m)}(n)
\end{align}
\subsubsection{The Array Output Case}

For a convolution layer, any given single weight value \(w_{z,i}(n)\) is used in the calculation of \(M=OH\texttt{x}OW\) output pixels. Therefore, in a single time step, \(w_{z,i}(n)\) has weights updates from M sources. Because the weight update contributions from each output pixel are summed together to create the total weight update for \(w_k(n)\), these updates are operationally independent of each other. Therefore, instead of a single optimization with M constraints, there are M separate optimization problems, each with a scalar Lagrangian multiplier. Therefore, to extend (\ref{eq:PMD_CNN_scalar}) over an array of output pixels, we sum over all constraint equations. Starting with (\ref{eq:PMD_CNN_scalar}), sum over M:

 \begin{align*}
    \sum_{m=0}^{M-1}w_{z,i}(n+1) & = \sum_{m=0}^{M-1}\Big[w_{z,i}(n) + \frac{\delta^{(m)}(n)}{||\mathbb{X}_i^{(m)}(n)||^2} x_{z,i}^{(m)}(n)\Big]
\end{align*}
 \begin{align}
    w_{z,i}(n+1) & = w_{z,i}(n) + \frac{1}{M}\sum_{m=0}^{M-1}\Big[\frac{\delta^{(m)}(n)}{||\mathbb{X}_i^{(m)}(n)||^2} x_{z,i}^{(m)}(n)\Big]
\end{align}
\subsubsection{Normalized Inputs}\label{PMD_Normalized_Inputs}

The explicit normalization by the input power in the weight update equation can be removed if the input is normalized to zero mean and unit variance. To do so, assume that within each feature map patch with dimensions HxW, the pixels locally have the same statistics and therefore have the same variance. Then the variance for \(x_{z,i}^{(m)}\) is the same for all \(m\) and \(z\), and is indicated by \(\sigma_i^2\). 

Let us introduce the learning parameter \(\mu\), and roll \(\frac{1}{M}\) into \(\mu\). Assume that the input along channel \(i\) is zero mean, denoted by \(u_i(n)\). Then \(\sigma_i^2\) can replace \(||\mathbb{X}_i^{(m)}(n)||^2\):

 \begin{align}
    w_{z,i}(n+1) = w_{z,i}(n) + \mu\sum_{m=0}^{M-1}\Big[\frac{\delta^{(m)}(n)}{\sigma_i^2} u_{z,i}^{(m)}(n)\Big]
\end{align}

\subsubsection{The Role of Batches}

In Section \ref{PMD_Normalized_Inputs}, the variance is estimated over a single input feature map. For many CNNs, the number of pixels can be as small as 7x7, making it difficult to estimate the variance. However, if weight updates are calculated across  \(B\) input batches, the channel variances are now estimated from a size of \(B\texttt{x}IH\texttt{x}IW\), and the channel statistics estimate becomes more accurate. In the limit of large \(B\), it is safe to replace \(u_{z,i}^{(m)}(n)\) with its normalized version, \(u_{z,i}^{*(m)}(n)\). Then \(\sigma^2 = 1\) and can be removed:

 \begin{align} \label{eq:unit_var_input_update_B_limit}
    w_{z,i}(n+1) 
    & = w_{z,i}(n) + \mu^*\sum_{b=0}^{B-1}\sum_{m=0}^{M-1}\Big[\delta^{(m,b)}(n)u_{z,i}^{*(m,b)}(n)\Big]
\end{align}

\subsection{BatchNorm and NLMS}

In this section, we analyze how BatchNorm deviates from the prior analysis. To do so, we must start with defining the location of BatchNorm and NLMS in the forward and backward pass of CNN operations.

\begin{figure}[t]
\centering
\begin{subfigure}{0.24\textwidth}
  \centering
  \includegraphics[width=0.79\linewidth]{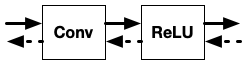}
  \caption{``Baseline''}
  \label{fig:NLMS_Cartoon_BasicCNN}
\end{subfigure}\hfil
\begin{subfigure}{0.24\textwidth}
  \centering
  \includegraphics[width=0.79\linewidth]{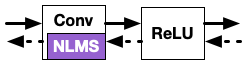}
  \caption{``NLMS\_L1'' and ``NLMS\_L2''}
  \label{fig:NLMS_Cartoon_CNN_NLMS}
\end{subfigure}\\
\begin{subfigure}{0.24\textwidth}
  \centering
  \includegraphics[width=1.0\linewidth]{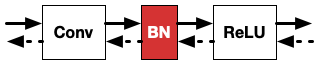}
  \caption{``BatchNorm''}
  \label{fig:NLMS_Cartoon_CNN_BN}
\end{subfigure}
\begin{subfigure}{0.24\textwidth}
  \centering
  \includegraphics[width=1.0\linewidth]{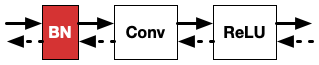}
  \caption{``BN\_Prior''}
  \label{fig:NLMS_Cartoon_BN_CNN}
\end{subfigure}\\
\caption{Layer arrangements. Solid arrows show the forward pass. Dashed arrows indicate the backward pass. NLMS is a layer that lies in the backward pass only.}
\label{fig:NLMS_Cartoon_Variants}
\end{figure}

\begin{figure*}[tbp]
\centering
\begin{subfigure}[b]{0.2\textwidth}
  \centering
  \includegraphics[width=1\linewidth]{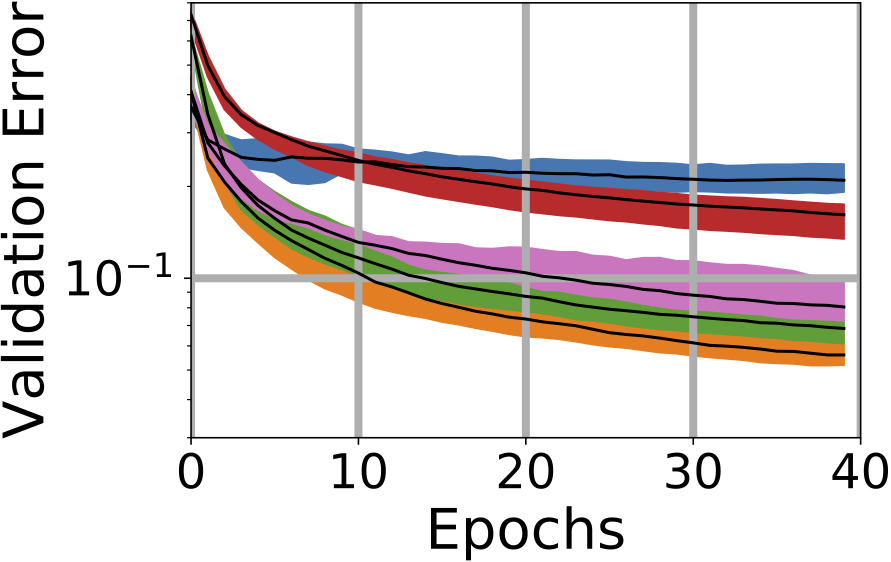}
  \caption{LR = 0.1 without noise}
  \label{fig:NLMS_0_1_NoNoise}
\end{subfigure}
\begin{subfigure}[b]{0.2\textwidth}
  \centering
  \includegraphics[width=1\linewidth]{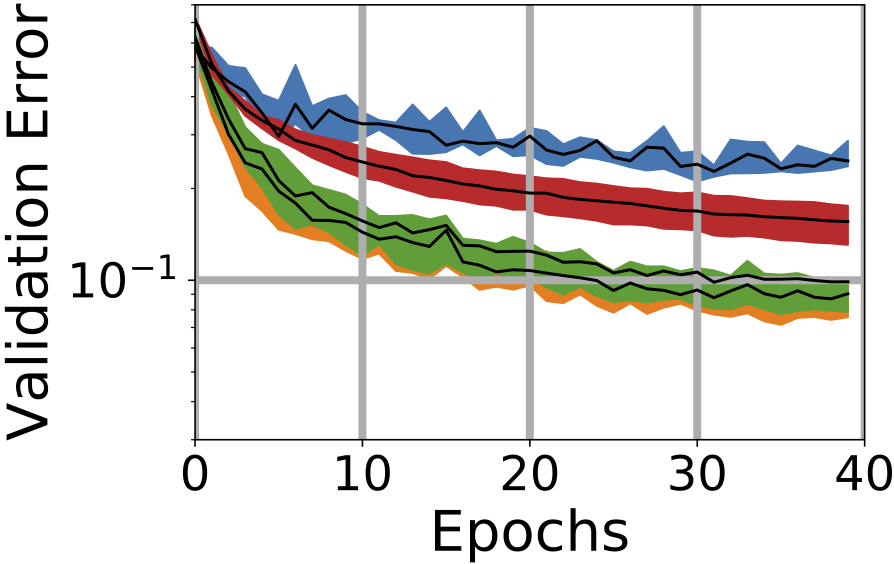}
  \caption{LR = 0.1 with noise}
  \label{fig:NLMS_0_1_WithNoise}
\end{subfigure}
\begin{subfigure}[b]{0.2\textwidth}
  \centering
  \includegraphics[width=1\linewidth]{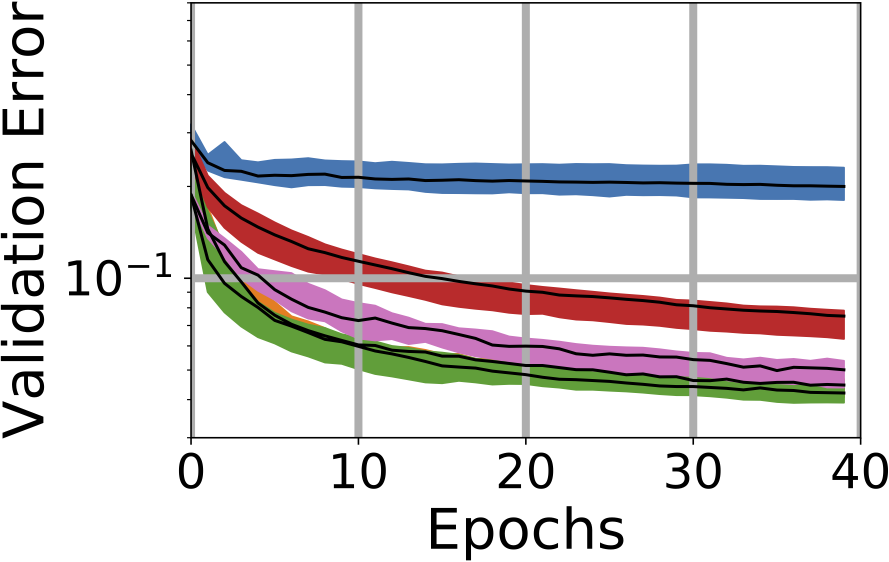}
  \caption{LR = 1.0 without noise}
  \label{fig:NLMS_1_0_NoNoise}
\end{subfigure}
\begin{subfigure}[b]{0.2\textwidth}
  \centering
  \includegraphics[width=1\linewidth]{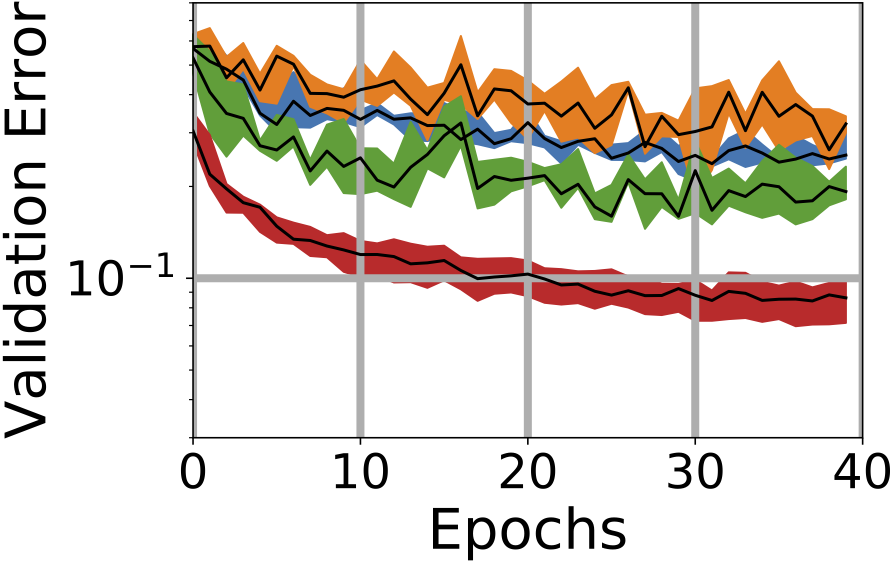}
  \caption{LR = 1.0 with noise}
  \label{fig:NLMS_1_0_WithNoise}
\end{subfigure}
\begin{subfigure}[b]{0.1\textwidth}
  \centering
  \includegraphics[width=1\linewidth]{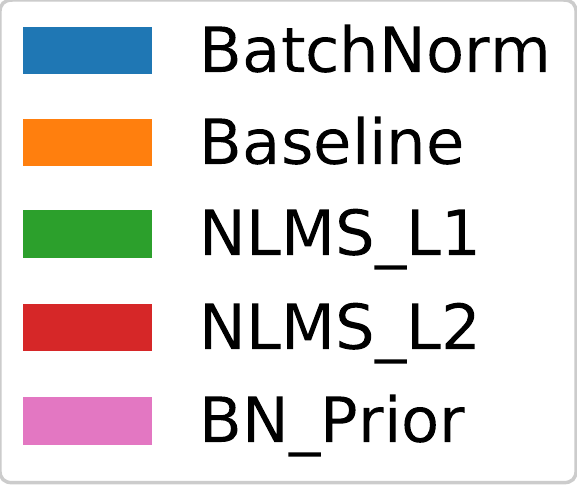}
  \label{fig:NLMS_legend}
\end{subfigure}
\caption{NLMS training curves over 40 epochs. Bands denote \nth{2} to \nth{3} quartile spread of the error.}
\label{fig:NLMS_1_0_Results}
\end{figure*}

\subsubsection{BatchNorm Before or After the Convolution}

Fig. \ref{fig:NLMS_Cartoon_BasicCNN} gives the base arrangement of a convolution layer followed by a ReLU nonlinearity. Fig. \ref{fig:NLMS_Cartoon_CNN_NLMS} illustrates how NLMS operates on the weight update during the backward pass. Fig. \ref{fig:NLMS_Cartoon_CNN_BN} and \ref{fig:NLMS_Cartoon_BN_CNN} show the two possible placements of BatchNorm, either before or after the convolution layer. It is not clear from \cite{ioffe2015batch} which placement of the BatchNorm layer is better. Work such as \cite{Mishkin2016} found that the networks performed slightly better when the BatchNorm layer is placed before the convolution layer. Despite these findings, the standard convention is to place BatchNorm after the convolution layer. 

We can only remove the explicit normalization term and still satisfy PMD if the inputs are already normalized. Assuming the learned parameters of BatchNorm are close to their initial values, BatchNorm placed before the convolution layer more closely satisfies PMD. From Fig. \ref{fig:NLMS_Cartoon_Variants}, it is clear that BatchNorm has no chance of normalizing the inputs of that particular layer if placed after a given convolution layer, and thus cannot meet the conditions outlined in Section \ref{PMD_Normalized_Inputs}. We do not expect it to match the performance of NLMS.

\subsubsection{The Role of Learned Parameters}
A key difference between BatchNorm and the normalization requirements of PMD are the BatchNorm channel-wise scale and shift parameters, \(\gamma\) and \(\beta\), respectively. These parameters change the channel-wise mean from 0 to \(\beta\) and the variance from 1 to \(\gamma^2\). 

To follow PMD exactly, the common CNN update equation needs the following change:

 \begin{align}
    w_{z,i}(n+1) & = w_{z,i}(n) + \mu^*\sum_{b=0}^{B-1}\sum_{m=0}^{M-1}\Big[\frac{\delta^{(m,b)}(n)}{(\gamma_i^2 +\beta_i^2)} x_{z,i}^{(m,b)}(n)\Big]
\end{align}
where \(\gamma\) and \(\beta\) are usually initialized to 1 and 0, respectively. This means that in the first training step, the CNN update follows PMD exactly. The first training step is arguably the point in training when NLMS is most needed. During the first training steps, when the weights have not settled, the input power is still wildly fluctuating. Recall that the original motivation behind NLMS was that it enables designers to pick a high \(\mu\) when the input power is strongly fluctuating.

Even though \(\gamma\) and \(\beta\) allow deviations from the weight update size mandated by PMD, as long as they do not deviate too far from their initial values, we expect BatchNorm placed before the convolution layer to come close to satisfying PMD.

\subsection{Experiments with NLMS}
Here, we quantify the effect of NLMS on gradient noise, and compare NLMS to BatchNorm before and after the convolution operation. The network variants (Fig. \ref{fig:NLMS_Cartoon_Variants}) are:

\begin{itemize}
    \item Baseline: Network in Table \ref{Tab:Baseline}
    \item BatchNorm: BatchNorm layer after the convolution
    \item NLMS\_L1: NLMS, with L1 norm
    \item NLMS\_L2: NLMS, with L2 norm
    \item BN\_Prior: BatchNorm layer before the convolution
\end{itemize}

To isolate the effects of NLMS on the convolution layers and not have the FC layers' learning power compensate for the impact of noise in the weight updates, we leave the FC layer as untrained. We only apply BatchNorm and NLMS to the second convolution layer. This results in BatchNorm having a higher validation error, dropping the baseline's accuracy from 99\% to 95\%. This is acceptable for our analysis because we are only interested in the validation error variance due to noise injection. To measure the effect of NLMS in a controlled manner, we inject noise into the local error (before weight gradient calculation). The noise is drawn from a Gaussian and scaled according to the local error variance. We train with stochastic gradient descent, no dropout, and no weight decay or momentum over 40 epochs (Fig. \ref{fig:NLMS_1_0_Results}). 

Fig. \ref{fig:NLMS_1_0_Results} shows that NLMS, which satisfies PMD exactly, has the least amount of noise amplification, resulting in the smoothest curve in the presence of noise. NLMS\_L2 has the smallest band and shows the greatest resilience to noise. BN\_Prior, which comes close to satisfying PMD, performs similar to NLMS\_L2. Other networks, including both BatchNorm and Baseline, are sensitive to injected noise. As a result of leaving the FC layer untrained, the variant with BatchNorm placed after the convolution layer settles at a higher error.

\section{Related Works}\label{section: Related Works}

Our work is similar to others that focus on the Hessian matrix because the input autocorrelation matrix \(\mathbb{R}\) is the expectation of the local Hessian of the layer weights. Reference \cite{Zhang2018} studies the relationship between the local Hessian and backpropagation and similarly proposed that BatchNorm applied to an FC layer controls the spectrum of the local Hessian eigenvalues, which can lead to better training speed. In this work, we study BatchNorm applied to convolution layers and separate the amplification and suppression effects of BatchNorm to demonstrate that it is the amplification of the smallest eigenvalues that leads to increased training speed. The works of \cite{Lecun1993}\cite{LeCun2012} derive the Hessian and draw conclusions similar to our work. They extend the findings to Hessian-free techniques for determining an adaptive learning rate. We derive the relationship between BatchNorm and the eigenvalues of \(\mathbb{R}\) by applying adaptive filter ideas (principle of orthogonality and Wiener-Hopf equations).

\section{Conclusion}
This work uses the lens of classical adaptive filters to analyze BatchNorm. We showed that CNN weight updates have natural modes, whose bounds control stability and training speed dynamics. We explored these connections with experiments on LeNet/MNIST and ResNet20/CIFAR10. For small LeNet-like networks, the experiments match the theoretical results. For ResNet, the stability behavior provides an explanation of how BatchNorm stabilizes networks with residual connections. The ResNet experiments demonstrate that amplification at low learning rates is critical for fast convergence. BatchNorm does not demonstrate similar amplification results at low learning rates, and this holds back training. Further work is needed to fully understand the role of BatchNorm in ResNets at low \(\mu\). Finally, we derived NLMS for CNNs and showed how BatchNorm placed before the convolution operation is closest to NLMS and demonstrates similar gradient noise suppression.

\bibliographystyle{ieeetr}
\bibliography{references}

\begin{thebibliography}{10}

\bibitem{He2016}
K.~He, X.~Zhang, S.~Ren, and J.~Sun, ``{Deep residual learning for image
  recognition},'' in {\em Proceedings of the IEEE Computer Society Conference
  on Computer Vision and Pattern Recognition}, pp.~770--778, 2016.

\bibitem{ioffe2015batch}
S.~Ioffe and C.~Szegedy, ``Batch normalization: Accelerating deep network
  training by reducing internal covariate shift,'' in {\em Proceedings of the
  32nd International Conference on Machine Learning}, pp.~448--456, 2015.

\bibitem{Hanin2018}
B.~Hanin, ``{Which neural net architectures give rise to exploding and
  vanishing gradients?},'' in {\em Advances in Neural Information Processing
  Systems 32}, pp.~582--591, 2018.

\bibitem{yang2019MeanFieldTheoryBatchNorm}
Y.~Greg, J.~Pennington, V.~Rao, J.~Sohl{-}Dickstein, and S.~Schoenholz, ``A
  mean field theory of batch normalization,'' in {\em 7th International
  Conference on Learning Representations}, 2019.

\bibitem{PyTorchNEURIPS2019_9015}
A.~Paszke, S.~Gross, F.~Massa, A.~Lerer, J.~Bradbury, G.~Chanan, T.~Killeen,
  Z.~Lin, N.~Gimelshein, L.~Antiga, A.~Desmaison, A.~Kopf, E.~Yang, Z.~DeVito,
  M.~Raison, A.~Tejani, S.~Chilamkurthy, B.~Steiner, L.~Fang, J.~Bai, and
  S.~Chintala, ``Pytorch: An imperative style, high-performance deep learning
  library,'' in {\em Advances in Neural Information Processing Systems 33},
  pp.~8024--8035, 2019.

\bibitem{santurkar2018does}
S.~Santurkar, D.~Tsipras, A.~Ilyas, and A.~Madry, ``How does batch
  normalization help optimization?,'' in {\em Advances in Neural Information
  Processing Systems 32}, pp.~2488--2498, 2018.

\bibitem{Zhang2019}
H.~Zhang, Y.~Dauphin, and T.~Ma, ``{Fixup initialization: Residual learning
  without normalization},'' in {\em 7th International Conference on Learning
  Representations}, 2019.

\bibitem{Balduzzi2017}
D.~Balduzzi, M.~Frean, L.~Leary, J.~Lewis, K.~Ma, and B.~McWilliams, ``{The
  shattered gradients problem: If resnets are the answer, then what is the
  question?},'' in {\em Proceedings of the 34th International Conference on
  Machine Learning}, pp.~536--549, 2017.

\bibitem{widrow1960adaline}
B.~Widrow, ``An adaptive adaline neuron using chemical memistors,'' Tech. Rep.
  TR-1553-2, Stanford Electronics Labs, Stanford, CA, Oct. 1960.

\bibitem{widrow1960adaptive}
B.~Widrow and M.~Hoff, ``Adaptive switching circuits,'' Tech. Rep. TR-1553-1,
  Stanford Electronics Labs, Stanford, CA, June 1960.

\bibitem{Haykin:2002}
S.~Haykin, {\em Adaptive Filter Theory}.
\newblock Upper Saddle River, NJ: Prentice Hall, 4th~ed., 2002.

\bibitem{widrow1971adaptive}
B.~Widrow, ``Adaptive filters,'' in {\em Aspects of Network and System Theory}
  (R.~Kalman and N.~DeClaris, eds.), pp.~563--587, New York: Holt, Rinehart,
  and Winston, 1971.

\bibitem{krizhevsky2009learning}
A.~Krizhevsky and G.~Hinton, ``Learning multiple layers of features from tiny
  images,'' 2009.

\bibitem{le1989handwritten}
Y.~LeCun, L.~Jackel, B.~Boser, J.~Denker, H.~Graf, I.~Guyon, D.~Henderson,
  R.~Howard, and W.~Hubbard, ``Handwritten digit recognition: Applications of
  neural network chips and automatic learning,'' {\em IEEE Communications
  Magazine}, vol.~27, no.~11, pp.~41--46, 1989.

\bibitem{lecun1998gradientMNIST}
Y.~LeCun, L.~Bottou, Y.~Bengio, and P.~Haffner, ``Gradient-based learning
  applied to document recognition,'' {\em Proceedings of the IEEE}, vol.~86,
  no.~11, pp.~2278--2324, 1998.

\bibitem{Krizhevsky2012AlexNet}
A.~Krizhevsky, I.~Sutskever, and G.~E. Hinton, ``Imagenet classification with
  deep convolutional neural networks,'' in {\em Advances in Neural Information
  Processing Systems 25}, (Red Hook, NY, USA), p.~1097–1105, Curran
  Associates Inc., 2012.

\bibitem{simonyan2014very}
K.~Simonyan and A.~Zisserman, ``Very deep convolutional networks for
  large-scale image recognition,'' {\em arXiv preprint arXiv:1409.1556}, 2014.

\bibitem{glorot2010understanding}
X.~Glorot and Y.~Bengio, ``Understanding the difficulty of training deep
  feedforward neural networks,'' in {\em Proceedings of the Thirteenth
  International Conference on Artificial Intelligence and Statistics},
  pp.~249--256, 2010.

\bibitem{he2015delving}
K.~He, X.~Zhang, S.~Ren, and J.~Sun, ``Delving deep into rectifiers: Surpassing
  human-level performance on imagenet classification,'' in {\em Proceedings of
  the IEEE International Conference on Computer Vision}, pp.~1026--1034, 2015.

\bibitem{Mishkin2016}
D.~Mishkin and J.~Matas, ``{All you need is a good init},'' in {\em 4th
  International Conference on Learning Representations}, pp.~1--13, 2016.

\bibitem{Zhang2018}
H.~Zhang, W.~Chen, and T.~Liu, ``On the local hessian in back-propagation,'' in
  {\em Advances in Neural Information Processing Systems 32}, pp.~6520--6530,
  2018.

\bibitem{Lecun1993}
Y.~LeCun, P.~Simard, and B.~Pearlmutter, ``Automatic learning rate maximization
  by on-line estimation of the hessian's eigenvectors,'' in {\em Advances in
  Neural Information Processing Systems 7}, pp.~156--163, 1993.

\bibitem{LeCun2012}
Y.~LeCun, L.~Bottou, G.~Orr, and K.~M{\"{u}}ller, ``{Efficient BackProp},'' in
  {\em Neural Networks: Tricks of the Trade}, pp.~9--48, Springer, 2012.

\end{thebibliography}
\end{document}